\documentclass[10pt,journal,compsoc]{IEEEtran}

\ifCLASSOPTIONcompsoc
  \usepackage[nocompress]{cite}
\else
  \usepackage{cite}
\fi

\usepackage [autostyle, english = american]{csquotes}
\usepackage{amsfonts}
\MakeOuterQuote{"}
\usepackage{amsmath,array,graphicx}
\usepackage{graphics} 
\usepackage{setspace}   
\usepackage[hyphens]{url}
\usepackage{hyperref}
\usepackage[]{subfig}
\usepackage{color}
\usepackage{multirow}
\graphicspath{{images/}}
\begin{document}

\title{Deep Learning for Vision-based Prediction: A Survey}
\author{Amir~Rasouli
\IEEEcompsocitemizethanks{\IEEEcompsocthanksitem A. Rasouli is with Noah's Ark Laboratory at Huawei Technologies Canada,
19 Allstate Pkwy, Markham, ON L3R 5A4 \protect\\
E-mail: amir.rasouli@huawei.com}
}


\IEEEtitleabstractindextext{%
\begin{abstract}
Vision-based prediction algorithms have a wide range of applications including autonomous driving, surveillance, human-robot interaction, weather prediction.  The objective of this paper is to provide an overview of the field in the past five years with a particular focus on deep learning approaches. For this purpose, we categorize these algorithms into video prediction, action prediction, trajectory prediction, body motion prediction, and other prediction applications. For each category, we highlight the common architectures, training methods and types of data used. In addition, we discuss the common evaluation metrics and datasets used for vision-based prediction tasks. A database of all the information presented in this survey, cross-referenced according to papers, datasets and metrics, can be found online at \url{https://github.com/aras62/vision-based-prediction}. 
\end{abstract}

\begin{IEEEkeywords}
Video Prediction, Action Prediction , Trajectory Prediction, Motion Prediction, Survey.
\end{IEEEkeywords}}

\maketitle

\IEEEdisplaynontitleabstractindextext
\IEEEpeerreviewmaketitle

\IEEEraisesectionheading{\section{Introduction}\label{sec:introduction}}

\IEEEPARstart{T}{he} ability to predict the changes in the environment and the behavior of objects is fundamental in many applications such as surveillance, autonomous driving, scene understanding, etc. Prediction is a widely studied field in various artificial intelligence communities. A subset of these algorithms relies primarily on visual appearances of the objects and the scene to reason about the future. Other approaches use different forms of sensors such as wearable or environmental sensors to learn about the past states of the environment or objects. 

The focus of this report is on vision-based prediction algorithms, which primarily use visual information to observe the changes in the environment and predict the future. In this context, prediction can be in the form of generating future scenes or reasoning about specific aspects of the objects, e.g. their trajectories, poses, etc.

For this review, we divide the prediction algorithms into five groups, namely video prediction, action prediction, trajectory prediction,  motion (pose) prediction, and others which involve various applications of prediction such as trend prediction, visual weather prediction, map prediction, semantic prediction, etc. In addition, we briefly discuss algorithms that use a form of prediction as an intermediate step to perform tasks such as object detection, action detection, and recognition, etc. Moreover, for each group of prediction algorithms, we will talk about the common datasets and metrics and discuss of their characteristics. It should be noted that due to the broad scope of this review and the large body of work on the vision-based prediction, this review will only focus on works that had been published since five years ago in major computer vision, robotics and machine learning venues.  In addition, as the title of the paper suggests, the main focus of the discussion will be on deep learning methods given their popularity in recent years.

\section{Vision-based Prediction}
Before reviewing the works on vision-based prediction algorithms, there are a number of points that should be considered.

\subsection{Applications}
Based on our review, we have identified four major vision-based applications namely, video prediction, action prediction, trajectory prediction, and  motion prediction. We discuss each of the studies in each category in a dedicated section. Some of the prediction works, such as visual weather prediction, semantic prediction, contests outcome prediction, that do not fit to any of the four major categories are presented in other application section. 

Some works address multiple prediction tasks, e.g. predicting trajectories and actions simultaneously, and therefore might fall in more than one category. It should be noted that we only include an algorithm in each category if the corresponding task is directly evaluated. For instance, if an algorithm performs video prediction for future action classification, and only evaluates the accuracy of predicted actions, it will only appear in the action prediction category. Furthermore, some works that are reviewed in this paper propose multiple architectures, e.g. recurrent and feedforward, for solving the same problem. In architecture-based categorizations, these algorithms may appear more than once.

\subsection{Methods}
\subsubsection{Algorithms}
This work focuses on vision-based algorithms, which use some form of visual input such as RGB camera images or active sensors such as LIDAR. It should be noted that many algorithms, especially trajectory prediction ones, only use ground truth data such as object trajectories without actual visual processing, e.g. for detection of objects. However, as long as these algorithms are evaluated on vision datasets, they are included in this paper. Note that a completer list of papers with published code can be found in Appendix \ref{papers_with_code}.

\subsubsection{Architectures}
As mentioned earlier, we focus on algorithms that have a deep learning component, either in the stage of visual representation generation (e.g. using convolutional features) or reasoning (e.g. using an MultiLayer Preceptron (MLP) for classification). We will, however, acknowledge the classical methods by mentioning some of the main techniques and including them in the datasets and metrics sections of this paper.

We classify the algorithms in terms of training techniques and architectures. In practice, this is very challenging as the majority of algorithms use a combination of different approaches. For example, recurrent networks often rely on a form of Convolutional Neural Networks (CNNs) to generate feature representations for scenes, poses of agents, etc. To better distinguish between different classes of algorithms, we only focus on the core component of each algorithm, i.e. the parts that are used for reasoning about the future. Hence, for example, if an algorithm uses a CNN model for pre-processing input data and a recurrent network for temporal reasoning, we consider this algorithm as recurrent. On the other hand, if the features are used with a fully connected network, we categorize this algorithm as feedforward or one-shot method. A few algorithms propose the use of both architectures for reasoning. We address those methods as hybrid. In addition, it should be noted that many works propose alternative approaches using each architecture. Therefore, we categorize them in more than one group.

\subsubsection{Data type}
As one would expect,  vision-based algorithms primarily rely on visual information. However, many algorithms use pre-trained off-the-shelf algorithms to transform the input to some explicit feature spaces, e.g. poses, trajectories, action labels and perform reasoning in those feature spaces. If pre-processing is not part of the main algorithm, we consider those secondary features as different types of data inputs to the algorithms. If some basic processing, e.g. generating convolutional features for a scene is used, we consider the data type of the original input, e.g. RGB images.

\section{Video prediction}
Video or future scene prediction can be regarded as the most generic form of prediction. The objective of video prediction algorithms is to generate future scenes, often in the form of RGB images \cite{Gao_2019_ICCV,Kwon_2019_CVPR, Castrejon_2019_ICCV, Ho_2019_ICCV} and/or optical flow maps \cite{Zhang_2019_ICIP, Liang_2017_ICCV,Gao_2019_ICCV, Ji_2018_WACV}. The generated images in turn can be used for various tasks such as action prediction \cite{Zeng_2017_ICCV}, event prediction \cite{Gujjar_2019_ICRA}, flow estimation \cite{Liang_2017_ICCV}, semantic segmentation \cite{Terwilliger_2019_WACV}, etc.

Video prediction applications rely on generative models whose task is to predict future scene(s) based on a short observation of input sequences (or in some cases only a single image \cite{Li_2018_ECCV}). Although many approaches use feedforard architectures \cite{Kwon_2019_CVPR, Gao_2019_ICCV, Ho_2019_ICCV, Zhang_2019_ICIP, Gujjar_2019_ICRA, Reda_2018_ECCV, Li_2018_ECCV,Ji_2018_WACV, Bhattacharjee_2018_ACCV, Ying_2018_ACCV,Zeng_2017_ICCV, Bhattacharjee_2017_NIPS}, the majority of algorithms take advantage of Recurrent Neural Networks (RNNs) such as Gated Recurrent Units (GRUs) \cite{Oliu_2018_ECCV}, Long-Short Term Memory (LSTM) networks \cite{Ye_2019_ICCV, Kim_2019_NIPS, Wang_2019_BMVC, Ho_2019_ICIP, Tang_2019_ICIP, Cai_2018_ECCV, Hsieh_2018_NIPS,Xu_2018_NIPS, Wichers_2018_ICML, Walker_2017_ICCV, Wang_2017_NIPS, Villegas_2017_ICML, Finn_2016_NIPS, Oh_2015_NIPS}, its variation Convolutional LSTMs (ConvLSTMs)\cite{Castrejon_2019_ICCV,Lee_2019_BMVC,Xu_2018_CVPR, Byeon_2018_ECCV,Liu_2018_ECCV,Jin_2018_IROS, Lu_2017_CVPR, Liang_2017_ICCV,Finn_2016_NIPS} or a combination of these \cite{Jung_2019_IROS, Wang_2017_NIPS}.

Generative Adversarial Networks (GANs) are particularly popular in the video prediction community. \cite{Kwon_2019_CVPR, Kim_2019_NIPS, Lee_2019_BMVC, Wang_2019_BMVC, Tang_2019_ICIP, Cai_2018_ECCV, Xu_2018_NIPS, Bhattacharjee_2018_ACCV, Ying_2018_ACCV, Lu_2017_CVPR, Liang_2017_ICCV, Walker_2017_ICCV, Bhattacharjee_2017_NIPS}. In these adversarial training frameworks, there are two compoents: A generative network that produces future representations and  a discriminator whose objective is to distinguish between the predicted representations (e.g. optical flow \cite{Liang_2017_ICCV}, frames \cite{Kim_2019_NIPS}, motion \cite{Lee_2019_BMVC}) or their temporal consistency \cite{Kwon_2019_CVPR, Wang_2019_BMVC} and the actual ground truth data by producing a binary classification score that indicates whether the prediction is real or fake. While many algorithms use discriminators to judge how realistic the final generated images \cite{Kim_2019_NIPS, Tang_2019_ICIP, Cai_2018_ECCV, Xu_2018_NIPS, Bhattacharjee_2018_ACCV, Lu_2017_CVPR} are or intermediate features (e.g. poses \cite{Walker_2017_ICCV}), others use multiple discriminators at different stages of processing.  For example, the authors of \cite{Kwon_2019_CVPR, Wang_2019_BMVC} use two discriminators, one is responsible for judging the temporal consistency of the generated frames (i.e. whether the order of generated frames is real) and the other assesses whether the generated frames are real or not. Lee et al. \cite{Lee_2019_BMVC} use three discriminators to assess the quality of generated frames and the intermediate motion and content features. Using a two-stream approach, the method in \cite{Liang_2017_ICCV} produces both the next frame and optical flow and each stream is trained with a separate discriminator. The prediction network of \cite{Bhattacharjee_2017_NIPS} uses a discriminator for intermediate features generated from input scenes and another discriminator for the final results.

Variational Autoencoders (VAEs) \cite{Kingma_2013_ARXIV} or Conditional VAEs (CVAEs) \cite{Rezende_2014_ARXIV} are also used in some approaches \cite{Castrejon_2019_ICCV, Kim_2019_NIPS, Li_2018_ECCV, Hsieh_2018_NIPS, Walker_2017_ICCV}. VAEs model uncertainty in generated future frames by defining a posterior distribution over some latent variable space \cite{Castrejon_2019_ICCV, Hsieh_2018_NIPS, Walker_2017_ICCV}. In CVAEs, the posterior is conditioned on an additional parameter such as the observed action in the scenes \cite{Kim_2019_NIPS} or initial observation \cite{Li_2018_ECCV}. Using VAEs, at inference time, a random sample is drawn from the posterior to generate the future frame.

Many video prediction algorithms operate solely on input images and propose various architectural innovations for encoding the content and generating future images \cite{Kwon_2019_CVPR, Castrejon_2019_ICCV, Xu_2018_CVPR, Byeon_2018_ECCV, Oliu_2018_ECCV, Liu_2018_ECCV, Bhattacharjee_2018_ACCV, Ying_2018_ACCV, Jin_2018_IROS, Zeng_2017_ICCV, Bhattacharjee_2017_NIPS, Wang_2017_NIPS}. For example, the method in \cite{Kwon_2019_CVPR} performs a two-way prediction, forward and backward. Each prediction relies on two discriminators for assessing the quality of the generated images and temporal consistency. The model presented in \cite{Castrejon_2019_ICCV} trains a context network by inputting an image sequence into a ConvLSTM whose output is used to initialize convolutional networks responsible for generating the next frames.  Xu et al.  \cite{Xu_2018_CVPR}, in addition to raw pixel values, encode the output of a high pass filter applied to the image as a means of maintaining the structural integrity of the objects in the scene. In \cite{Bhattacharjee_2017_NIPS}, the authors use a two-step approach in which they first perform a coarse frame prediction followed by a fine frame prediction. In \cite{Bhattacharjee_2018_ACCV}, the algorithm learns in two stages. A discriminator is applied after features are generated from the scenes and another one after the final generated frames. 

Optical flow prediction has been widely used as an intermediate step in video prediction algorithms \cite{Gao_2019_ICCV, Ho_2019_ICCV, Lee_2019_BMVC, Ho_2019_ICIP, Zhang_2019_ICIP, Li_2018_ECCV, Wichers_2018_ICML, Lu_2017_CVPR, Liang_2017_ICCV}. For example, to deal with occlusion in dynamic scenes, Gao et al. \cite{Gao_2019_ICCV} disentangle flow and pixel-level predictions into two steps: the algorithm first predicts the flow of the scene, and then uses it, in conjunction with the input frames, to predict the future. Similar multi-step approaches have also been used in \cite{Liang_2017_ICCV, Lu_2017_CVPR,Li_2018_ECCV, Lee_2019_BMVC}. In \cite{Lee_2019_BMVC}, the authors use two separate branches: one branch receives two consecutive frames $(t, t+1)$ and produces context information. The second branch produces motion information by receiving two frames that are $k$ steps apart (i.e. $t+1$, $t+k$). The outputs of these two branches are fused and fed into the final scene generator.  The method in \cite{Liang_2017_ICCV} simultaneously produces the next future frame and the corresponding optical flow map. In this architecture, two additional networks are used: A flow estimator which uses the output of the frame generator and the last observation to estimate a flow map and a warping layer which performs differential 2D spatial transformation to warp the last observed image into the future predicted frame according to the predicted flow map. 

Some algorithms rely on various intermediate steps for video prediction\cite{Ye_2019_ICCV, Kim_2019_NIPS, Tang_2019_ICIP, Cai_2018_ECCV, Hsieh_2018_NIPS, Xu_2018_NIPS, Ji_2018_WACV, Ying_2018_ACCV, Jin_2018_IROS}. For instance, the method in \cite{Ye_2019_ICCV}, reasons about the locations and features of individual entities (e.g. cubes) for final scene predictions.  Kim et al. \cite{Kim_2019_NIPS} first identify keypoints, which may correspond to important structures such as joints, and then predict their motion. For videos involving humans, in \cite{Tang_2019_ICIP, Cai_2018_ECCV, Ji_2018_WACV} the authors identify and reason about the changes in poses, and use this information to generate future frames. In \cite{Ying_2018_ACCV, Jin_2018_IROS}, in addition to raw input images, the differences between consecutive frames used in the learning process.

Prediction networks can also be provided with additional information to guide future frame generation. In \cite{Wang_2019_BMVC, Reda_2018_ECCV} an optical flow network and in \cite{Walker_2017_ICCV, Villegas_2017_ICML} a pose estimation network are used in addition to RGB images. Using a CVAE architecture, in \cite{Kim_2019_NIPS} the authors use the action lables as conditional input for frame generation. In the context of active tasks, e.g. object manipulation with robotic arms, in which the consequences of actions influence the future scene configuration, it is common to condition the future scene generation on the current or future intended actions \cite{Jung_2019_IROS, Finn_2016_NIPS, Oh_2015_NIPS}.

\subsection{Summary}
Video prediction algorithms are based on generative models that produce future images given a short observation, or in extreme cases a single view of the scene. Both recurrent and feedforward models are widely used in the field, with recurrent ones being slightly more favorable. The architectural designs and training strategies such as the VAEs or GANs are very common. However, it is hard to establish which one of these approaches is superior given that the majority of the video prediction algorithms are application-agnostic, meaning that they are evaluated on a wide range of video datasets with very different characteristics such as traffic scenes, activities, games, object manipulations, etc. (more on this in Section \ref{datasets}).

Despite the great progress in the field, video prediction algorithms are still facing some major challenges. One of them is the ability to hallucinate, which is to generate visual representations for parts of the scenes that were not visible during observation phase, e.g. due to occlusions. This is particularly an issue for more complex images such as traffic scenes, movies, etc. The complexity of the scenes also determines how fast the generated images would degrade. Although these algorithms show promising results in simple synthetic videos or action sequences, they still struggle in real practical applications. In addition, many of these algorithms cannot reason about the expected presence or absence of objects in the future. For example, if a moving object is present in the observations and is about to exit the field of view in near future, the algorithms account for it in the future scenes as long as parts of it are visible in the observation stage. This can be an issue for safety-critical applications such as autonomous driving in which the presence or absence of traffic elements and the interactions between them are essential for action planning. 


\section{Action prediction}
Action prediction algorithms can be categorized into two groups: Next action or event prediction (or action anticipation) and early action prediction. In the former category, the algorithms use the observation of current activities or scene configurations and predict what will happen next. Early action prediction algorithms, on the other hand, observe parts of the current action in progress and predict what this action is. The classical learning approaches such as Conditional Random Fields (CRFs) \cite{Schulz_2015_IV}, Support Vector Machines (SVMs) with hand-crafted features \cite{Hu_2016_ECCV, Schneemann_2016_IROS, Volz_2016_ITSC, Xu_2015_ICCV, Zhang_2015_ICRA, Kohler_2015_ITSC, Volz_2015_ITSC}, Markov models \cite{Luo_2019_IROS, Wu_2019_IROS, Rhinehart_2017_ICCV, Kwak_2017_IPT, Hu_2016_IROS, Jain_2015_ICCV}, Bayesian networks \cite{Hariyono_2016_IES, Hashimoto_2015_ICVES} and other statistical methods \cite{Joo_2019_CVPR, Ziaeetabar_2018_IROS, Qi_2017_ICCV, Park_2016_IROS, Mahmud_2016_ICIP, Xu_2016_ICIP, Arpino_2015_ICRA} have been widely used in recent years. However, as mentioned earlier, we will only focus on deep learning approaches.

\subsection{Action anticipation}
Action prediction algorithms are used in a wide range of applications including cooking activities  \cite{Ke_2019_CVPR, Furnari_2019_ICCV, Sener_2019_ICCV, Gammulle_2019_BMVC, Alati_2019_ICIP, Furnari_2019_ICIP, Farha_2018_CVPR, Mahmud_2017_ICCV}, traffic understanding \cite{Rasouli_2019_BMVC, Ding_2019_ICRA, Gujjar_2019_ICRA, Saleh_2019_ICRA, Scheel_2019_ICRA, Aliakbarian_2018_ACCV, Scheel_2019_ICRA, Strickland_2018_ICRA, Casas_2018_CORL, Rasouli_2017_ICCVW, Chan_2016_ACCV, Jain_2016_ICRA}, accident prediction \cite{Manglik_2019_IROS, Wang_2019_ICIP, Suzuki_2018_CVPR, Zeng_2017_CVPR}, sports \cite{Su_2017_CVPR, Felsen_2017_ICCV} and other forms of activities \cite{Liang_2019_CVPR, Sun_2019_CVPR, Shen_2018_ECCV, Schydlo_2018_ICRA_2, Zhong_2018_ICIP, Zeng_2017_ICCV, Gao_2017_BMVC, Vondrick_2016_CVPR_2, Kataoka_2016_BMVC, Zhou_2015_ICCV}. Although the majority of these algorithms use sequences in which the objects and agents are fully observable, a number of methods rely on egocentric scenes \cite{Ke_2019_CVPR, Furnari_2019_ICCV, Furnari_2019_ICIP, Shen_2018_ECCV, Su_2017_CVPR, Zhou_2015_ICCV} which are recorded from the point of view of the acting agents and only parts of their bodies (e.g. hands) are observable.

Action prediction methods predominantly use a variation of RNN-based architectures including LSTMs \cite{Liang_2019_CVPR, Furnari_2019_ICCV, Sener_2019_ICCV, Gammulle_2019_BMVC, Scheel_2019_ICRA, Alati_2019_ICIP, Furnari_2019_ICIP, Wang_2019_ICIP, Shen_2018_ECCV, Aliakbarian_2018_ACCV, Scheel_2019_ICRA, Schydlo_2018_ICRA_2, Zhong_2018_ICIP, Su_2017_CVPR, Zeng_2017_CVPR, Felsen_2017_ICCV, Mahmud_2017_ICCV, Gao_2017_BMVC, Jain_2016_CVPR, Chan_2016_ACCV, Jain_2016_ICRA}, GRUs \cite{Sun_2019_CVPR, Rasouli_2019_BMVC, Ding_2019_ICRA, Farha_2018_CVPR}, ConvLSTMs \cite{Strickland_2018_ICRA}, and Quasi-RNNs (QRNNs) \cite{Suzuki_2018_CVPR}. For instance, in \cite{Sun_2019_CVPR, Alati_2019_ICIP} the authors use a graph-based RNN architecture in which the nodes represent actions and the edges of the graph represent the transitions between the actions. The method in \cite{Farha_2018_CVPR} employs a two-step approach: using a recognition algorithm, the observed actions and their durations are recognized. These form a one-hot encoding vector which is fed into GRUs for the prediction of the future activities, their corresponding start time and length. In the context of vehicle behavior prediction, Ding et al. \cite{Ding_2019_ICRA} uses a two-stream GRU-based architecture to encode the trajectory of two vehicles and a shared activation unit to encode the vehicles mutual interactions. Scheel et al. \cite{Scheel_2018_ICRA} encode the relationship between the ego-vehicle and surrounding vehicles in terms of their mutual distances.  The vectorized encoding is then fed into a bi-directional LSTM. At each time step, the output of the LSTM is classified, using a softmax activation, into a binary value indicating whether it is safe for the ego-vehicle to change lane. In \cite{Suzuki_2018_CVPR} the authors use a QRNN network to capture the relationships between road users in order to predict the likelihood of a traffic accident. To train the model, the authors propose an adaptive loss function that assigns penalty weights depending on how early the model can predict accidents. 

As an alternative to recurrent architectures, some algorithms use feedforward architectures using both 3D \cite{Ke_2019_CVPR, Gujjar_2019_ICRA, Saleh_2019_ICRA} and 2D \cite{Manglik_2019_IROS, Farha_2018_CVPR, Casas_2018_CORL, Felsen_2017_ICCV, Rasouli_2017_ICCVW, Zeng_2017_ICCV, Vondrick_2016_CVPR_2, Kataoka_2016_BMVC, Zhou_2015_ICCV} convolutional networks. For example, in the context of pedestrian crossing prediction, in \cite{Gujjar_2019_ICRA} the authors use a generative 3D CNN model that produces future scenes and is followed by a classifier.  The method of \cite{Saleh_2019_ICRA} detects and tracks pedestrians in the scenes, and then feeds the visual representations of the tracks, in the form of an image sequence, into a 3D CNN architecture, which directly classifies how likely the pedestrian will cross the road. To predict the time of traffic accidents, the method in \cite{Manglik_2019_IROS} processes each input image using a 2D CNN model and then combines the representations followed by a fully-conntected (\textit{fc}) layer for prediction. Farha et al. \cite{Farha_2018_CVPR} create a 2D matrix by stacking one-hot encodings of actions for each segment of observation and use a 2D convolutional net to generate future actions encodings. Casas et al. \cite{Casas_2018_CORL} use a two-stream 2D CNN, each processing the stacked voxelized LIDAR scans and the scene map. The feature maps obtained from each stream are fused and fed into a backbone network followed by three headers responsible for the detection of the vehicles and predicting their intentions and trajectories. For sports forecasting, Felsen et al. \cite{Felsen_2017_ICCV}  concatenate 5 image observations channel-wise and feed the resulting output into a 2D CNN network comprised of 4 convolutional layers and an \textit{fc} layer.

Although some algorithms rely on a single source of information, e.g. a set of pre-processed features from RGB images \cite{Sun_2019_CVPR, Wang_2019_ICIP, Suzuki_2018_CVPR, Zhong_2018_ICIP, Zeng_2017_CVPR, Felsen_2017_ICCV, Mahmud_2017_ICCV, Gao_2017_BMVC, Jain_2016_CVPR} or trajectories \cite{Ding_2019_ICRA}, many algorithms use a multimodal approach by using various sources of information such as optical flow maps \cite{Furnari_2019_ICCV, Alati_2019_ICIP, Furnari_2019_ICIP, Aliakbarian_2018_ACCV, Zhou_2015_ICCV}, poses \cite{Liang_2019_CVPR, Rasouli_2019_BMVC, Alati_2019_ICIP, Schydlo_2018_ICRA_2, Jain_2016_ICRA}, scene attributes (e.g. road structure, semantics) \cite{Liang_2019_CVPR, Scheel_2019_ICRA, Shen_2018_ECCV, Casas_2018_CORL}, text \cite{Sener_2019_ICCV}, action labels \cite{Gammulle_2019_BMVC}, length of actions \cite{Farha_2018_CVPR}, speed (e.g. ego-vehicle or surrounding agents) \cite{Rasouli_2019_BMVC, Scheel_2019_ICRA, Aliakbarian_2018_ACCV, Scheel_2018_ICRA, Su_2017_CVPR}, gaze \cite{Shen_2018_ECCV, Schydlo_2018_ICRA_2}, current activities \cite{Rasouli_2017_ICCVW} and the time \cite{Ke_2019_CVPR} of the actions. For example, the method in \cite{Sener_2019_ICCV} uses a multi-stream LSTM in which two LSTMs encode visual features and cooking recipes and an LSTM decodes them for final predictions. To capture the relationships within and between sequences,  Gammulle et al. \cite{Gammulle_2019_BMVC} propose a two-stream LSTM network with external neural memory units. Each stream is responsible for encoding visual features and action labels. In \cite{Rasouli_2019_BMVC}, a multi-layer GRU structure is used in which features with different modalities enter the network at different levels and are fused with the previous level encodings. The fusion process is taking place according to the complexity of the data modality, e.g. more complex features such as encodings of pedestrian appearances enter the network at the bottom layer, whereas location and speed features enter at the second-last and last layers respectively. Farha et al. \cite{Farha_2018_CVPR} use a two-layer stacked GRU architecture which receives as input a feature tuple of the length of the activity and its corresponding one-hot vector encoding. In \cite{Aliakbarian_2018_ACCV}, the method uses a two-stage architecture: First information regarding the appearance of the scene, optical flow (pre-processed using a CNN) and vehicle dynamics are fed into individual LSTM units. Then, the output of these units is combined and passed through an \textit{fc} layer to create a representation of the context. This representation is used by another LSTM network to predict future traffic actions. In the context of human-robot interaction, the authors of \cite{Schydlo_2018_ICRA_2} combine the information regarding the gaze and pose of the humans using an encoder-decoder LSTM architecture to predict their next actions. Jain et al. \cite{Jain_2016_ICRA} use a fusion network to combine head pose information of the driver, outside scene features, GPS information, and vehicle dynamics to predict the driver's next action.

Before concluding this section, it is important to discuss the use of attention modules which have gained popularity in recent years \cite{Ke_2019_CVPR, Liang_2019_CVPR, Sun_2019_CVPR, Furnari_2019_ICCV, Scheel_2019_ICRA, Furnari_2019_ICIP, Wang_2019_ICIP, Shen_2018_ECCV, Zeng_2017_CVPR, Chan_2016_ACCV}. As the name implies, the objective of attention modules is to determine what has to be given more importance at a given time. These modules can come in different froms and can be applied to different dimensions of data and at various processing. Some of these modules are temporal attention \cite{Ke_2019_CVPR, Wang_2019_ICIP, Shen_2018_ECCV} for identifying keyframes, modality attention \cite{Furnari_2019_ICCV, Furnari_2019_ICIP} to prioritize between different modalities of data input, spatial attention \cite{Zeng_2017_CVPR, Chan_2016_ACCV} for highlighting the important parts of the scenes, and graph attention \cite{Sun_2019_CVPR} for weighting nodes of the graph. In some cases, a combination of different attention mechanisms is used  \cite{Liang_2019_CVPR, Scheel_2019_ICRA}.

\subsubsection{Summary}
In the field of action anticipation, RNN architectures are strongly preferred. Compared to feedforward algorithms, recurrent methods have the flexibility of dealing with variable observation lengths and multi-modal data, in particular, when they are significantly different, e.g. trajectories and RGB images. However, basic recurrent architectures such as LSTMs and GRUs rely on some forms of pre-processing, especially when dealing with high dimensional data such as RGB images, which requires the use of various convolutional networks, a process that can be computationally costly. Feedforward models, on the other hand, can perform prediction in one shot, meaning that they can simultaneously perform temporal reasoning and spatial feature generation in a single framework, and as a result, potentially have a shorter processing time. 

Many of the approaches mentioned earlier are generative in nature. They generate  representations in some feature space and then using these representations predict what will happen next. Some algorithms go one step further and generate the actual future images and use them for prediction. Although such an approach seems effective for single actor events, e.g. cooking scenes, human-robot interaction, it is not a feasible approach for multi-agent predictions such as reasoning about behaviors of pedestrians or cars in traffic scenes.
   
The majority of the methods reviewed in this section use multi-modal data input. This seems to be a very effective approach, especially in high dimensional problems such as predicting road user behavior where the state (e.g. location and speed) of the road user, observer, and other agents, as well as scene structure, lighting conditions, and many other factors, play a role in predicting the future behavior.

Multi-tasking, e.g. predicting actions and trajectories, are an effective way to predict future actions. For instance, trajectories can imply the possibility of certain actions, e.g. moving towards the road implies the possibility that the person might cross the street. As a result, the simultaneous learning of different tasks can be beneficial. 

Last but not least is the use of attention modules. These modules are deemed to be very effective , in particular for tasks with high complexity in terms of the modality of input data, the scene structure and temporal relations. 

\subsection{Early action prediction}
Similar to action anticipation methods, early action prediction algorithms widely use recurrent network architectures \cite{Sun_2019_CVPR, Wang_2019_CVPR, Gammulle_2019_ICCV, Zhao_2019_ICCV, Yao_2018_CVPR, Shi_2018_ECCV, Butepage_2018_ICRA, Cho_2018_WACV, Aliakbarian_2017_ICCV, Li_2016_WACV}. Although many of these algorithms use similar aproaches to action anticipation algorithms, some propose new approaches. For example, in \cite{Wang_2019_CVPR} the authors use a teacher-student learning scheme where the teacher learns to recognize full sequences using a bi-directional LSTM and the student relies on partial videos using an LSTM network. They perform knowledge distillation by linking feature representations of both networks. Using GAN frameworks, the methods in \cite{Gammulle_2019_ICCV, Shi_2018_ECCV} predict future feature representations of videos in order to predict actions. Zhao et al. \cite{Zhao_2019_ICCV} implement a Kalman filter using an LSTM architecture. In this method, actions are predcted after observing each frame and corrections are made if the next observation provide additional information. The method of \cite{Aliakbarian_2017_ICCV} uses a two-step LSTM architecture which first generates an encoding of the context using context-aware convolutional features and then combines these encodings with action-aware convolutional features to predict the action. The authors of this method propose a new loss function that penalizes false negatives with the same strength at any point in time and false positives with a strength that increases linearly over time, to reach the same weight as that on false negatives. In \cite{Li_2016_WACV} the authors perform coarse-to-fine-grained predictions depending on the amount of evidence available for the type of action. 

Many early action prediction methods adopt feedforward architectures \cite{Safaei_2019_WACV, Liu_2018_CVPR_ssnet, Chen_2018_ECCV, Cho_2018_WACV, Butepage_2017_CVPR, Kong_2017_CVPR, Singh_2017_ICCV, Lee_2016_ICPR}. The authors of \cite{Safaei_2019_WACV} predict actions from a single image by transforming it into a new representation called Ranked Saliency Map and Predicted Optical Flow.  This representation is passed through a 2D convent and \textit{fc} layers for final prediction. In \cite{Liu_2018_CVPR_ssnet, Cho_2018_WACV}, the authors use temporal CNN (TCNNs) architectures, which are a series of 1D dilated convolutional layers designed to capture the temporal dependencies of feature representations, e.g. in the form of a vector representation of poses \cite{Liu_2018_CVPR_ssnet} or word-embeddings \cite{Cho_2018_WACV} describing the video frames. Chen et al. \cite{Chen_2018_ECCV} use features of body parts generated by a CNN model and train an attention module whose objective is to activate only features that contribute to the prediction of the action. In \cite{Singh_2017_ICCV}, the authors use an action detection framework, which incrementally predicts the locations of the actors and the action classes based on the current detections.
 
The majority of the early action prediction algorithms pre-process the entire observed scenes using, e.g. different forms of convolutional neural networks \cite{Wang_2019_CVPR, Zhao_2019_ICCV, Safaei_2019_WACV, Shi_2018_ECCV, Cho_2018_WACV, Kong_2017_CVPR, Aliakbarian_2017_ICCV, Lee_2016_ICPR} or other forms of features \cite{Li_2016_WACV}. Some algorithms complement these features using optical flow maps \cite{Gammulle_2019_ICCV, Cho_2018_WACV, Singh_2017_ICCV}. Another group of action prediction methods focuses on specific parts of the observations. For example, in \cite{Liu_2018_CVPR_ssnet, Yao_2018_CVPR, Butepage_2018_ICRA, Butepage_2017_CVPR} the authors use the changes in the poses of actors to predict their actions. The method in \cite{Chen_2018_ECCV} uses body parts extracted by cropping a local patch around identified joints of the actors. The authors of \cite{Lee_2016_ICPR} only use the visual appearances of actors extracted from detected bounding boxes, and the relationship between them.

\subsubsection{Summary}
Early action detection methods have many commonalities with action anticipation algorithms in terms of architectural design. However, there are some exceptions that are more applicable in this domain. These exceptions are teacher-student training schemes for knowledge distillation, identifying discriminative features, and recursive prediction/correction mechanisms. In addition, early action prediction algorithm, with a few exceptions, rely on single modal data for prediction and rarely use refinement frameworks such as attention modules. Adopting these techniques and operating on multi-modal feature spaces can further improve the performance of early action prediction algorithms. Unlike the action anticipation methods, there is no strong preference for recurrent or feedforward approaches. Some approaches take advantage of architectures such as temporal CNNs which are popular in the language processing domain and show their effectiveness for early action prediction tasks. 

\section{Trajectory prediction}
As the name implies, trajectory prediction algorithms predict the future trajectories of objects, i.e. the future positions of the objects over time. These approaches are particularly popular for applications such as intelligent driving and surveillance. Predicted trajectories can be used directly, e.g. in route planning for autonomous vehicles, or used for predicting future events, anomalies, or actions. 

In this section, we follow the same routine as before and focus on algorithms that have a deep learning component while acknowledging many recent works that have used classical approaches including Gaussian mixture models \cite{Carvalho_2019_IROS, Yoo_2016_CVPR} and processes \cite{Mogelmose_2015_IV, Zhou_2016_ICRA}, Markov decision processes (MDPs) \cite{Rudenko_2018_ICRA, Rudenko_2018_IROS, Schulz_2018_IROS, Shen_2018_IROS, Shkurti_2017_IROS, Vasquez_2016_ICRA, Karasev_2016_ICRA, Lee_2016_WACV, Bai_2015_ICRA}, Markov chains \cite{Solaimanpour_2017_ICRA, Chen_2016_ICRA} and other techniques \cite{Sanchez_2019_ICIP, Lee_2018_ICRA, Hasan_2018_WACV, Ballan_2016_ECCV, Pfeiffer_2016_IROS, Vo_2015_ICRA, Akbarzadeh_2015_IROS, Schulz_2015_ITSC}.

Trajectory prediction applications like many other sequence prediction tasks heavily rely on recurrent architectures such as LSTMs \cite{Chang_2019_CVPR, Liang_2019_CVPR, Sadeghian_2019_CVPR, Zhang_2019_CVPR, Zhao_2019_CVPR, Bi_2019_ICCV, Choi_2019_ICCV, Huang_2019_ICCV, Rasouli_2019_ICCV, Thiede_2019_ICCV, Kosaraju_2019_NIPS, Ding_2019_ICRA_2, Li_2019_ICRA, Anderson_2019_IROS, Srikanth_2019_IROS, Zhu_2019_IROS, Xue_2019_WACV, Gupta_2018_CVPR, Hasan_2018_CVPR, Xu_2018_CVPR_encoding, Yao_2018_CVPR,  Pfeiffer_2018_ICRA, Rehder_2018_ICRA, Xue_2018_WACV, Bartoli_2018_ICPR, Alahi_2016_CVPR}, and GRUs  \cite{Hong_2019_CVPR, Rhinehart_2019_ICCV, Li_2019_IROS, Fernando_2018_ACCV, Lee_2017_CVPR, Rhinehart_2018_ECCV}. These methods often use an encoder-decoder architecture in which a network, e.g. an LSTM encodes single- or multi-modal observations of the scenes for some time, and another network generates future trajectories given the encoding of the observations. Depending on the complexity of input data, these algorithms may rely on some form of pre-processing for generating features or embedding mechanisms to minimize the dimensionality of the data.

The feedforward algorithms  \cite{Hong_2019_CVPR, Kim_2019_ICCV, Chai_2019_CORL, Cui_2019_ICRA, Huang_2019_ICRA, Zhou_2019_IROS, Jain_2019_CORL, Baumann_2018_ICRA, Casas_2018_CORL, Zhang_2018_CORL, Ma_2017_CVPR, Yi_2016_ECCV} often use whole views of the scenes (i.e. the environment and moving objects) and encode them using convolutional layers followed by a regression layer to predict trajectories. A few algorithms use hybrid approaches in which both convolutional and recurrent reasoning are also used \cite{Chandra_2019_CVPR, Li_2019_CVPR}.  

Depending on the prediction task, algorithms may rely on single- or multi-modal observations. For example, in the context of visual surveillance where a fixed camera provide a top-down or bird-eye viewing angle, many algorithms only use past trajectories of the agents in either actual 2D frame coordinates or their velocities calculated by the changes from each time step to another \cite{Li_2019_CVPR, Zhang_2019_CVPR, Huang_2019_ICCV, Thiede_2019_ICCV, Anderson_2019_IROS, Zhu_2019_IROS, Zhi_2019_CORL, Gupta_2018_CVPR, Xu_2018_CVPR_encoding, Fernando_2018_ACCV, Pfeiffer_2018_ICRA, Vemula_2018_ICRA, Xue_2019_WACV, Bartoli_2018_ICPR, Alahi_2016_CVPR}. In addition to observations of individual trajectories of agents, these algorithms focus on modeling the interaction between the agents and how they impact each other. For example, Zhang et al. \cite{Zhang_2019_CVPR} use a state refinement module that aligns all pedestrians in the scene with a message passing mechanism that receives as input the current locations of the subjects and their encodings from an LSTM unit. In \cite{Huang_2019_ICCV} a graph-based approach is used where pedestrians are considered as nodes and the interactions between them as edges of the graph. By aggregating information from neighboring nodes, the network learns to assign a different level of importance to each node for a given subject. The authors of  \cite{Gupta_2018_CVPR, Alahi_2016_CVPR} perform a pooling operation on the generated representations by sharing the state of individual LSTMs that have spatial proximity. 

As shown in some works, other sources of information are used in surveilling objects \cite{Liang_2019_CVPR, Sadeghian_2019_CVPR, Zhao_2019_CVPR, Choi_2019_ICCV, Kosaraju_2019_NIPS, Li_2019_IROS, Hasan_2018_CVPR, Yao_2018_CVPR, Pfeiffer_2018_ICRA, Xue_2018_WACV, Lee_2017_CVPR, Ma_2017_CVPR, Yi_2016_ECCV}. For example, in addition to encoding the interactions with the environment, Liang et al. \cite{Liang_2019_CVPR} use the semantic information of the scene as well as changes in the poses of the pedestrians. In \cite{Sadeghian_2019_CVPR, Zhao_2019_CVPR, Choi_2019_ICCV, Kosaraju_2019_NIPS, Li_2019_IROS, Xue_2018_WACV, Lee_2017_CVPR, Yi_2016_ECCV} the visual representations of the layout of the environment and the appearances of the subjects are included. The authors of \cite{Pfeiffer_2018_ICRA} use an occupancy map which highlights the potential traversable locations for the subjects. The method in \cite{Hasan_2018_CVPR} takes into account pedestrians' head orientations to estimate their fields of view in order to predict which subjects would potentially interact with one another. To predict interactions between humans, in \cite{Yao_2018_CVPR} the authors use both poses and trajectories of the agents. Ma et al. \cite{Ma_2017_CVPR}  go one step further and take into account the pedestrians' characteristics (e.g. age, gender) within a game-theoretic perspective to determine how the trajectory of one pedestrians impact each other.

In the context of traffic understanding, predicting trajectories can be more challenging due to the fact that there is camera ego-motion involved (e.g. the prediction is from the perspective of a moving vehicle), there are interactions between different types of objects (e.g. vehicles and pedestrians), and there are certain constraints involved such as traffic rules, signals, etc. To achieve robustness, many methods in this domain take advantage of multi-modal data for trajectory prediction \cite{Chandra_2019_CVPR, Chang_2019_CVPR, Hong_2019_CVPR, Bi_2019_ICCV, Rasouli_2019_ICCV, Rhinehart_2019_ICCV, Cui_2019_ICRA, Ding_2019_ICRA_2, Huang_2019_ICRA, Tang_2019_ICRA, Cho_2019_IROS, Li_2019_IROS, Srikanth_2019_IROS, Chai_2019_CORL, Jain_2019_CORL, Bhattacharyya_2018_CVPR, Rhinehart_2018_ECCV, Baumann_2018_ICRA, Casas_2018_CORL, Lee_2017_CVPR}. In addition to using past trajectories, all these algorithms account for the road structure (whether it is from the perspective of the ego-vehicle or a top-down view) often in the form of raw visual inputs or, in some cases, as an occupancy map \cite{Bi_2019_ICCV, Baumann_2018_ICRA}. The scene layout can implicitly capture the structure of the road, the appearances of the objects (e.g. shape) and the dynamics (e.g. velocity or locations of subjects). Such implicit information can be further augmented by explicit data such as the shapes of the objects (in the case of vehicles) \cite{Chandra_2019_CVPR}, the speed \cite{Rasouli_2019_ICCV,Huang_2019_ICRA, Bhattacharyya_2018_CVPR} and steering angle \cite{Huang_2019_ICRA,Bhattacharyya_2018_CVPR} of the ego-vehicle, the distance between the objects \cite{Chang_2019_CVPR,Cho_2019_IROS}, traffic rules \cite{Cho_2019_IROS} and signals \cite{Casas_2018_CORL}, and kinematic constraints \cite{Zhang_2018_CORL}. For example, the method in \cite{Bhattacharyya_2018_CVPR} uses a two-stream LSTM encoder-decoder scheme: first stream encodes the current ego-vehicle's odometry (steering angle and speed) and the last observation of the scene and predicts future odometry of the vehicle. The second stream is a trajectory stream that jointly encodes location information of pedestrians and the ego-vehicle's odometry and then combines the encoding with the prediction of the odometry stream to predict the future trajectories of the pedestrians. The method in \cite{Rasouli_2019_ICCV} further extends this approach and adds an intention prediction stream which outputs how likely the observed pedestrian intends to cross the street. The intention likelihood is produced using an LSTM network that encodes the dynamics of the pedestrian, their appearances and their surroundings. Chandra et al. \cite{Chandra_2019_CVPR} create embeddings of contextual information by taking into account the shape and velocity of the road users and their spatial coordinates within a neighboring region. These embeddings are then fed into some LSTM networks followed by a number of convolutional layers to capture the dynamics of the scenes. In \cite{Bi_2019_ICCV} the authors use separate LSTMs for encoding the trajectories of pedestrians and vehicles (as orientated bounding boxes) and then combine them into a unified framework by generating an occupancy map of the scene centered at each agent, followed by a pooling operation to capture the interactions between different subjects. Lee et al. \cite{Lee_2017_CVPR} predict the future trajectories of vehicles in two steps: First, an encoder-decoder GRU architecture predicts future trajectories by observing the past ones. Then a refinement network adjusts the predicted trajectories by taking into account the contextual information in the forms of social interactions, dynamics of the agents involved, and the road structure.

Similar to action prediction algorithms, attention modules have been widely used in trajectory prediction methods \cite{Liang_2019_CVPR, Sadeghian_2019_CVPR, Zhang_2019_CVPR, Huang_2019_ICCV, Rasouli_2019_ICCV, Kosaraju_2019_NIPS, Li_2019_IROS, Xue_2019_WACV, Vemula_2018_ICRA, Fernando_2018_ACCV}. For example, in  \cite{Liang_2019_CVPR, Huang_2019_ICCV}, the attention module jointly measures spatial and temporal interactions. The authors of \cite{Sadeghian_2019_CVPR, Zhang_2019_CVPR, Kosaraju_2019_NIPS,Vemula_2018_ICRA} propose the use of social attention modules which estimate the relative importance of interactions between the subject of interest and its neighboring subjects. The method in  \cite{Rasouli_2019_ICCV} uses two attention mechanisms, a temporal attention module that measures the importance of each timestep of the observed trajectories and a series of self-attention modules which are applied to encodings of observations prior to predictions. Xue et al.  \cite{Xue_2019_WACV} propose an attention mechanism to measure the relative importance between different data modalities, namely the locations and velocities of subjects.

One of the major challenges in trajectory prediction is to model uncertainty of predictions, especially in scenarios where many possibilities exist (e.g. pedestrians at intersections).  Some algorithms such as \cite{Liang_2019_CVPR, Zhang_2019_CVPR, Rasouli_2019_ICCV, Xue_2019_WACV, Gupta_2018_CVPR, Alahi_2016_CVPR} train the models by directly measuring the error between the ground truth trajectories and predicted ones, e.g. by using an \textit{L2} objective function. At the inference time, these algorithms generate deterministic predictions. To measure the uncertainty of predictions, these models are trained multiple times with randomly initialized parameters. Alternatively, uncertainty can be estimated via probabilistic objective functions, e.g. Gaussian log-likelihood, as in  \cite{Chandra_2019_CVPR, Hong_2019_CVPR, Jain_2019_CORL, Choi_2019_ICCV, Hasan_2018_CVPR, Bhattacharyya_2018_CVPR} Instead of a single point in space, these algorithms predict a distribution that captures the uncertainty of the predictions. VAEs are another technique that can be used to estimate uncertainty \cite{Hong_2019_CVPR, Cho_2019_IROS, Li_2019_IROS, Felsen_2018_ECCV,Lee_2017_CVPR}. Using these methods, at training time a posterior distribution over some latent space $z$ is learned by conditioning, for example, over future trajectories. At the inference time, a random sample is drawn from the latent space for the predictions.

Since trajectory prediction algorithms are generative by nature, many approaches rely on adversarial training techniques \cite{Li_2019_CVPR, Sadeghian_2019_CVPR,Zhao_2019_CVPR, Thiede_2019_ICCV, Li_2019_ICRA, Anderson_2019_IROS, Li_2019_IROS, Gupta_2018_CVPR, Fernando_2018_ACCV} in which at training time a discriminator is used to predict whether the generated trajectories are real or fake. Kosaraju et al. \cite{Kosaraju_2019_NIPS} extend this approach by using two discriminators: A local discriminator which predicts the results on the output of the prediction using only past trajectories, and one global discriminator which operates on the output of the entire network, i.e. the prediction results based on trajectories and scene information.

\subsection{Summary}
Trajectory prediction is a widely studied field in the computer vision community. Although these works dominantly use recurrent network architectures, many approaches, such as those used in the field of traffic scene understanding, use feedforward networks. Trajectory prediction algorithms rely on one or more sources of information such as the past trajectories of subjects, surrounding visual context, object attributes, vehicle sensor readings, etc. One factor that is common in many of trajectory prediction algorithms is modeling the interactions between dynamic or dynamic and static objects. Relationships are captured explicitly or implicitly via encoding the scenes as a whole. Like many other prediction approaches, trajectory prediction algorithms benefit from various forms of attention mechanisms to learn the importance of spatial, temporal or social interactions between objects. To model uncertainty, techniques such as probabilistic objectives and variational encodings are used. 

Trajectory prediction algorithms are predominantly rely on past trajectory information to predict the future. Although past motion observations are very informative, in some context, e.g. traffic scenes, they are simply not enough. There is a need for a more explicit encoding of contextual information such as road conditions, the subject's attributes, rules and constraints, scene structure, etc. A number of approaches successfully have included a subset of these factors, but a more comprehensive approach should be considered.

\section{Motion prediction}
Although the term "motion prediction" in many cases is used to refer to future trajectory prediction, here we only consider the algorithms that are designed to predict changes in human pose. Motion prediction play a  fundamental role in all prediction approaches as an intermediate step, e.g. to reflect how the future visual representations would look like or the types of actions to anticipate. Like many other prediction applications, this field is dominated by deep learning models, even though some methods still rely on classical techniques \cite{Joo_2019_CVPR, Luo_2019_IROS, Cao_2015_WACV}.

Similar to other prediction algorithms, motion prediction methods widely use recurrent architectures such as LSTMs \cite{Yuan_2019_ICCV, Chiu_2019_WACV, Wu_2019_WACV, Yao_2018_CVPR,  Butepage_2018_ICRA, Chao_2017_CVPR, Walker_2017_ICCV} and GRUs \cite{Gopalakrishnan_2019_CVPR, Gui_2018_ECCV, Gui_2018_ECCV_2, Gui_2018_IROS, Martinez_2017_CVPR, Jain_2016_CVPR, Fragkiadaki_2015_ICCV} or a combination of both \cite{Wang_2019_ICCV}. For example, in \cite{Gopalakrishnan_2019_CVPR} the authors use a two-layer GRU model in which the top layer operates backward to learn noise processes and the bottom level is used to predict the poses given the past pose observations and the output of the top layer. Chiu et al. \cite{Chiu_2019_WACV} propose a hierarchical LSTM architecture in which each layer of the network encodes the observed poses at different time-scales. In the context of 3D pose prediction, some algorithms rely on a two-stage process where the visual inputs, either as a single image \cite{Chao_2017_CVPR} or a sequence of images \cite{Wu_2019_WACV}, are fed into a recurrent network to predict 2D poses of the agent. This is followed by a refinement procedure that transforms the 2D poses into 3D. 

Some approaches adopt feedforward architectures \cite{Mao_2019_ICCV, Hernandez_2019_ICCV, Zhang_2019_ICCV, Talignani_2019_IROS, Butepage_2017_CVPR}. For example, the method in  \cite{Hernandez_2019_ICCV} uses two feedforward networks in a two-stage process. First, the input poses are fed into an autoencoder which is comprised of fully connected layerss (implemented by 1D convolutions with a kernel size of 1) and self-attention blocks. The encodings are then used by multi-level 2D convolutional blocks for final predictions. Zhang et al. \cite{Zhang_2019_ICCV} predict 3D poses from RGB videos. In their method, the images are converted to a feature space using a convolutional network, and then the features are used to learn a latent 3D representation of 3D human dynamics. The representation is used by a network to predict future 3D poses. To capture movement patterns, a method proposed in \cite{Mao_2019_ICCV} converts poses into a trajectory space using discrete cosine transformation. The newly formed representations are then used in a Graph-CNN framework to learn the dependencies between different joint trajectories. 

To train motion prediction models, some authors use adversarial training methods in which a discriminator is used to classify whether the predicted poses are real or fake \cite{Hernandez_2019_ICCV, Gui_2018_IROS}. The discrimination procedure can also be applied to evaluating the continuity, i.e. correct order of, predictions as demonstrated in \cite{Gui_2018_ECCV_2}. In \cite{Wang_2019_ICCV} the discrimination score is used to generate a policy for future action predictions in the context of imitation learning.

\subsection{Summary}
Motion prediction algorithms primarily focus on the prediction of changes in the dynamics (i.e. poses) of observed agents. Such predictions can be fundamental to many other applications such as video or trajectory prediction tasks some of which were discussed previously. 

In recent works, recurrent network architectures are strongly preferred. The architecture of the choice often depends on the representation of the input data, e.g. whether joint coordinates are directly used or are encoded into a high-dimensional representation. 

Despite the development of many successful motion prediction algorithms, the majority of these methods rely on a single source of information, for example, poses or scenes. Encoding higher-level contextual information, such as scene semantics, interactions, etc. can potentially result in more robust predictions, as shown in other prediction applications. Attention modules also, except for one instance, haven't been adopted within motion prediction algorithms. Given the success of using attention in other prediction applications, motion prediction algorithms may benefit from the use of attention mechanisms.

\section{Other applications}
In the context of autonomous robotics, some algorithms are designed to predict occupancy grid maps (OGMs) that are grayscale representations of the robot's surroundings showing which parts of the environment are traversable.  These approaches are often object-agnostic and are concerned with generating future OGMs which are used by an autonomous agent to perform path planning. In recent years both classical \cite{Guizilini_2019_ICRA, Graves_2019_IROS, Afolabi_2018_IROS, Wilson_2015_ICRA} and deep learning \cite{Mohajerin_2019_CVPR, Katyal_2019_ICRA, Schreiber_2019_ICRA, Hoermann_2018_ICRA, Choi_2016_IROS} methods are used. The deep learning approaches, in essence, are similar to video prediction methods in which the model receives as input a sequence of OGMs and predicts the future ones over some period of time. In this context both recurrent \cite{Mohajerin_2019_CVPR, Schreiber_2019_ICRA, Choi_2016_IROS} and feedforward \cite{Katyal_2019_ICRA, Hoermann_2018_ICRA} methods were common. Another group of generative approaches is semantic map prediction algorithms \cite{Terwilliger_2019_WACV, Luc_2018_ECCV, Luc_2017_ICCV, Jin_2017_NIPS}. These algorithms receive as inputs RGB images of the scenes and predict future segmentation maps.

Some of the other vision-based prediction applications include weather \cite{Kim_2019_WACV, Chu_2018_WACV} and Solar irradiance forecasting \cite{Siddiqui_2019_WACV}, steering angle prediction \cite{Jin_2017_NIPS},  predicting the popularities of tweets based on  tweeted images used and the users' histories \cite{Wang_2018_WACV}, forecasting fashion trends \cite{Al-Halah_2017_ICCV}, storyline prediction \cite{Zeng_2017_ICCV}, pain anticipation \cite{Sur_2017_IROS}, predicting the effect of force after manipulating objects \cite{Mottaghi_2016_ECCV}, forecasting the winner of Miss Universe given the appearances of contestants' gowns \cite{Carvajal_2016_ICPR}, and predicting election results given the facial attributes of candidates \cite{Joo_2015_ICCV}. These algorithms rely on a combination of techniques discussed earlier in this paper. 

\section{Prediction in other vision applications}
Before concluding our discussion on vision-prediction methods, it is worth mentioning that prediction techniques are also widely used in other visual processing tasks such as video summarization  \cite{Lal_2019_WACV}, anomaly detection \cite{Liu_2018_CVPR}, tracking \cite{Fernando_2018_WACV}, active object recognition \cite{Jayaraman_2016_ECCV}, action detection \cite{Dave_2017_CVPR, Yang_2017_BMVC} and recognition \cite{Ziaeefard_2015_BMVC}. For example, tracking algorithms are very closely related to trajectory prediction ones and often rely on short term predictions to deal with gaps, e.g. due to occlusions, in tracking. For example,  in \cite{Fernando_2018_WACV} the method uses a recurrent framework to generate future frames in order to localize pedestrians in next frames. In the context of action detection, some methods rely on a future frame \cite{Dave_2017_CVPR} or trajectory prediction of objects to detect actions \cite{Yang_2017_BMVC}. In \cite{Jayaraman_2016_ECCV}, a method is used for detecting an object in 3D by relying on predicting next best viewing angle of the object. Liu et al. \cite{Liu_2018_CVPR} uses a video prediction framework to predict future motion flow maps and images. The future predictions that do not conform with expectations will be identified as abnormal.  

\section{The evaluation of state-of-the-art}
When it comes to the evaluation of algorithms, there are two important factors: metrics and datasets. They  highlight the strengths and weaknesses of the algorithms and provide a means to compare the relative performances of the methods. Given the importance of these two factors in the design of algorithms, we dedicate the following sections to discussing the common metrics and datasets used for vision-based prediction tasks. Since the datasets and metrics used in these applications are highly diverse, we will focus our discussion on some of the main ones for each prediction category while providing visual aids to summarize what the past works used for evaluation purposes. 

\section{Metrics}
\begin{figure*}
\centering
    \includegraphics[width=0.21\textwidth]{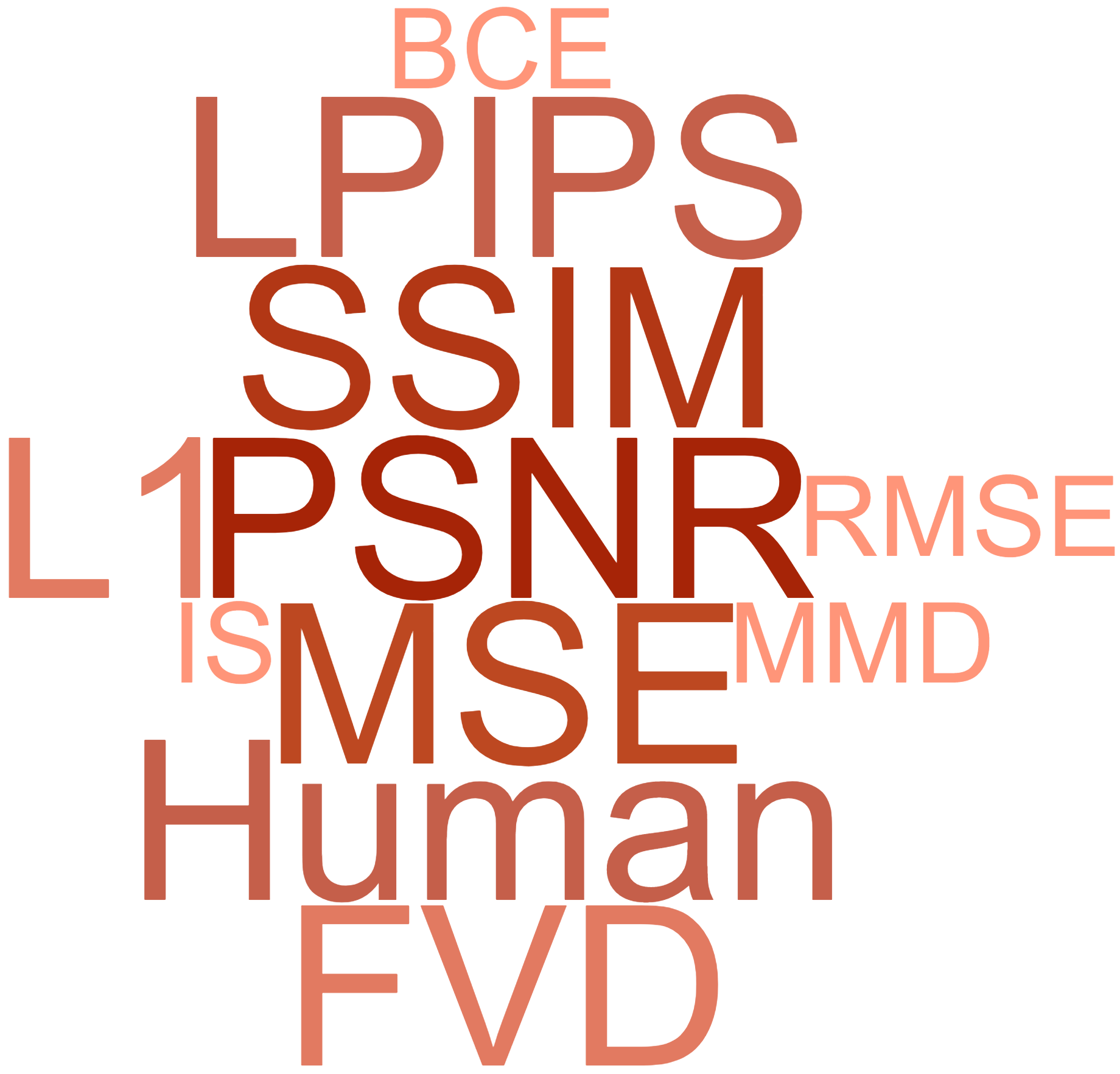}
    \hspace{0.1cm}
    \includegraphics[width=0.21\textwidth]{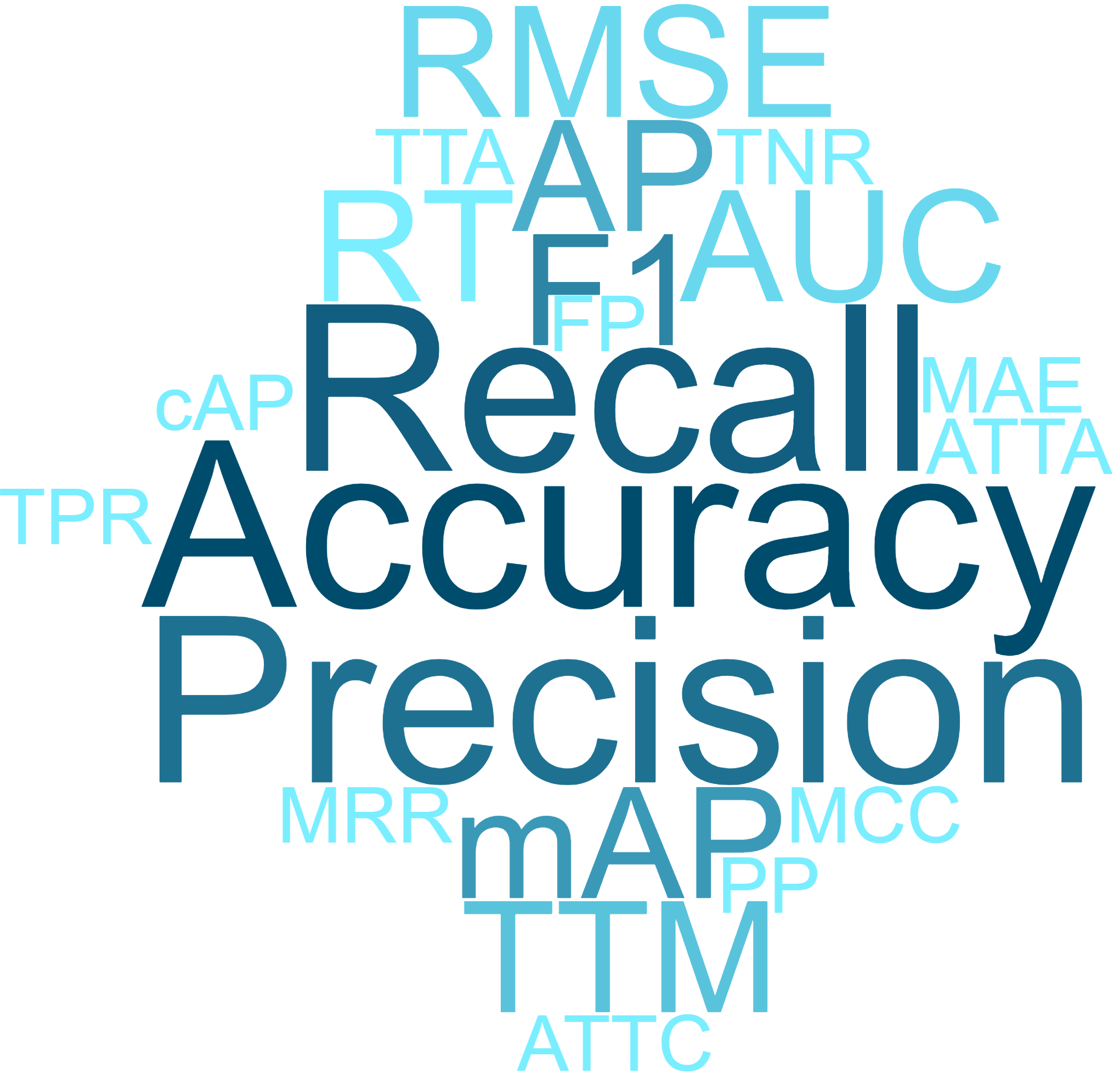}
    \hspace{0.1cm}
    \includegraphics[width=0.22\textwidth]{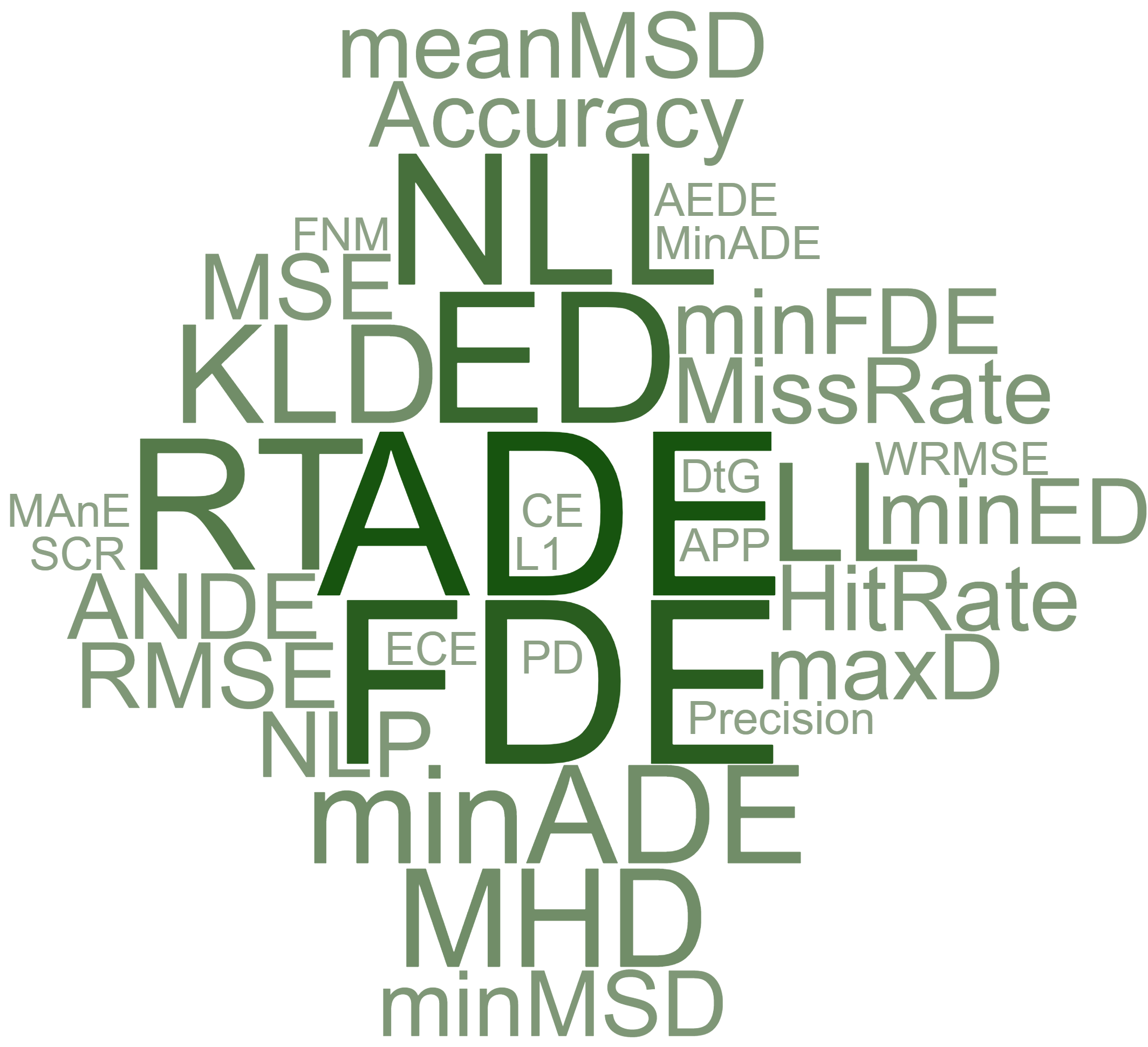}
    \hspace{0.1cm}
   \includegraphics[width=0.18\textwidth]{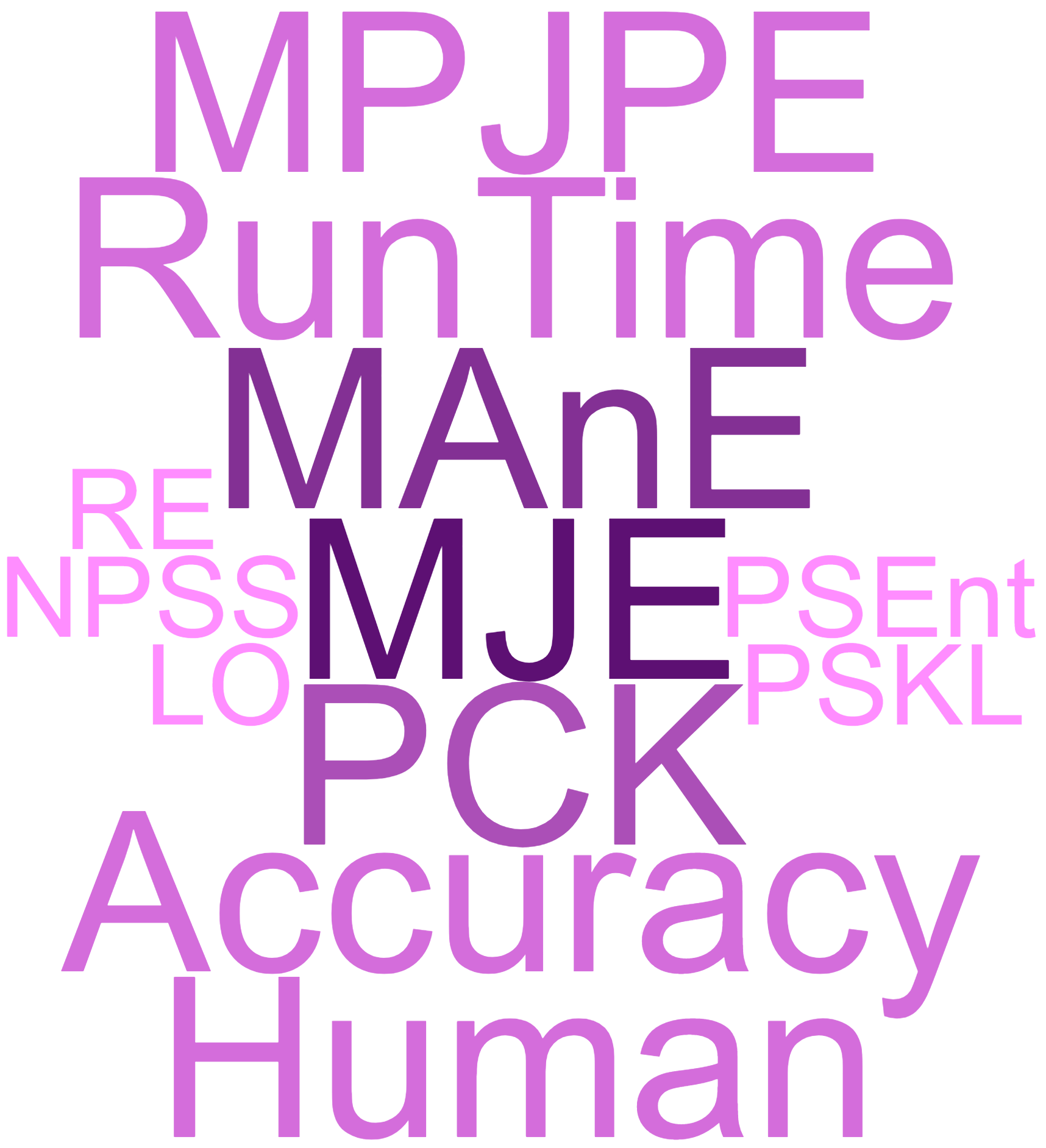}
   \caption{Metrics used in vision-based prediction applications. From left to right: Video, action, trajectory and motion prediction. }
\label{metrics_fig}
\end{figure*}

In this section, we follow the same routine as the discussion on the past works and divide the metrics into different categories. A summary of the metrics can be found in Figure \ref{metrics_fig}. The interested readers can also refer to Appendix \ref{metrics_papers} for further information regarding the metrics and the the papers that used them.  Note while we discuss the metrics in each category and we only provide mathematical expressions of the most popular metrics in Appendix \ref{appendix_metrics}.

\subsection{Video prediction}
Video prediction is about generating realistic images, hence the best performance is achieved when the disparities between the generated images and groundtruth images are minimal. The most straightforward way of computing the disparity is to measure pixel-wise error using a Mean Square Error (MSE) \cite{Kwon_2019_CVPR, Ho_2019_ICCV, Ho_2019_ICIP, Liu_2018_ECCV, Oliu_2018_ECCV, Reda_2018_ECCV, Zhao_2018_ECCV, Hsieh_2018_NIPS, Ying_2018_ACCV, Ji_2018_WACV, Liang_2017_ICCV, Wang_2017_NIPS, Oh_2015_NIPS}, which computes average squared intensity differences between pixels. Another more popular metric related to MSE is Peak Signal-to-Noise Ratio (PSNR) \cite{Kwon_2019_CVPR, Gao_2019_ICCV, Ho_2019_ICCV, Lee_2019_BMVC, Wang_2019_BMVC, Ho_2019_ICIP, Tang_2019_ICIP, Zhang_2019_ICIP, Xu_2018_CVPR, Byeon_2018_ECCV, Cai_2018_ECCV, Liu_2018_ECCV, Oliu_2018_ECCV, Reda_2018_ECCV, Zhao_2018_ECCV, Xu_2018_NIPS, Bhattacharjee_2018_ACCV, Ying_2018_ACCV, Jin_2018_IROS, Lu_2017_CVPR, Liang_2017_ICCV, Bhattacharjee_2017_NIPS, Wang_2017_NIPS, Villegas_2017_ICML, Finn_2016_NIPS}. PSNR is the ratio of the maximum pixel value (i.e. possible signal power), e.g. 255 in 8-bit images, divided by the MSE (or power of distorting noise) measure of two images. The lower the error between two images, the higher the value of PSNR, and consequently, the higher the quality of the generated images. Because of the wide dynamic range of signals, PSNR value is expressed in the logarithmic decibel scale. 

Although MSE, and PSNR metrics are easy to calculate, they cannot measure the perceived visual quality of a generated image. An alternative metric to address this issue is Structural SIMilarity (SSIM) index (\cite{Wang_2004_TIP}) \cite{Kwon_2019_CVPR, Castrejon_2019_ICCV, Gao_2019_ICCV, Ho_2019_ICCV, Lee_2019_BMVC, Wang_2019_BMVC, Ho_2019_ICIP, Tang_2019_ICIP, Zhang_2019_ICIP, Xu_2018_CVPR, Byeon_2018_ECCV, Cai_2018_ECCV, Liu_2018_ECCV, Oliu_2018_ECCV, Reda_2018_ECCV, Xu_2018_NIPS, Bhattacharjee_2018_ACCV, Ying_2018_ACCV, Jin_2018_IROS,  Liang_2017_ICCV, Bhattacharjee_2017_NIPS, Wang_2017_NIPS, Finn_2016_NIPS}  which is designed to model image distortion. To capture the structural differences between the two images, SSIM separates illumination information as it is independent of objects' structures. As a result, the similarity is measured by a combination of three comparisons, namely luminance, contrast, and structure. 

Higher-level contextual similarities may not be captured by distance measures on pixel values. More recently proposed metric, Learned Perceptual Image Patch Similarity (LPIPS), (\cite{Zhang_2018_CVPR})\cite{Castrejon_2019_ICCV, Ye_2019_ICCV, Jung_2019_IROS, Li_2018_ECCV}  measures the similarity between two images by comparing internal activations of convolutional networks trained for high-level classification tasks. The value is calculated by an average L2 distance over normalized deep features.

Some other metrics that have been used in the literature are qualitative human judgment \cite{Li_2018_ECCV, Wichers_2018_ICML, Zeng_2017_ICCV, Villegas_2017_ICML}, Frechet Video Distance (FVD) (\cite{Unterthiner_2018_arxiv}) \cite{Castrejon_2019_ICCV, Kim_2019_NIPS}, Maximum Mean Discrepancy (MMD) \cite{Walker_2017_ICCV}, Inception Scores (IS) (\cite{Salimans_2016_NIPS}) \cite{Walker_2017_ICCV}, Binary Cross Entropy (BCE) \cite{Hsieh_2018_NIPS}, L1 \cite{Gujjar_2019_ICRA, Reda_2018_ECCV}, and Root MSE (RMSE) \cite{Li_2018_ECCV}.  

\subsection{Action prediction}
Similar to classification tasks, many action prediction algorithms use accuracy measure to report on the performance  that is the ratio of the correct predictions with respect to the total number of predictions \cite{Joo_2019_CVPR, Ke_2019_CVPR, Furnari_2019_ICCV, Gammulle_2019_ICCV, Zhao_2019_ICCV, Gammulle_2019_BMVC, Rasouli_2019_BMVC, Gujjar_2019_ICRA, Scheel_2019_ICRA, Luo_2019_IROS, Wu_2019_IROS, Safaei_2019_WACV, Alati_2019_ICIP, Furnari_2019_ICIP, Farha_2018_CVPR, Liu_2018_CVPR_ssnet, Suzuki_2018_CVPR, Yao_2018_CVPR, Chen_2018_ECCV, Shen_2018_ECCV, Shi_2018_ECCV, Aliakbarian_2018_ACCV, Butepage_2018_ICRA, Scheel_2018_ICRA, Strickland_2018_ICRA, Zhong_2018_ICIP, Butepage_2017_CVPR, Kong_2017_CVPR, Su_2017_CVPR, Aliakbarian_2017_ICCV, Felsen_2017_ICCV, Mahmud_2017_ICCV, Rhinehart_2017_ICCV, Singh_2017_ICCV, Zeng_2017_ICCV, Cho_2018_WACV, Vondrick_2016_CVPR_2, Hu_2016_ECCV, Kataoka_2016_BMVC, Hu_2016_IROS, Park_2016_IROS, Schneemann_2016_IROS, Li_2016_WACV, Xu_2016_ICIP, Lee_2016_ICPR, Hariyono_2016_IES, Volz_2016_ITSC, Xu_2015_ICCV, Zhou_2015_ICCV, Arpino_2015_ICRA, Zhang_2015_ICRA, Hashimoto_2015_ICVES, Kohler_2015_ITSC, Schulz_2015_IV, Volz_2015_ITSC}. Despite being used widely, accuracy on its own is not a strong indicator of performance, especially when we are dealing with class-imbalance data. This is because, for example, the model can simply favor the more represented class and predict every input as that class. This then would result in a high accuracy measure because the metric only considers the ratio of correct predictions. To address these shortcomings, some works use complimentary metrics that, in addition to correct predictions, account for different types of false predictions. These metrics are precision \cite{Ke_2019_CVPR, Rasouli_2019_BMVC, Ding_2019_ICRA, Gujjar_2019_ICRA, Scheel_2019_ICRA, Alati_2019_ICIP, Wang_2019_ICIP, Suzuki_2018_CVPR, Mahmud_2017_ICCV, Qi_2017_ICCV, Kwak_2017_IPT, Jain_2016_CVPR, Jain_2016_ICRA, Hu_2016_IROS, Schneemann_2016_IROS, Jain_2015_ICCV}, recall \cite{Ke_2019_CVPR, Furnari_2019_ICCV, Sener_2019_ICCV, Rasouli_2019_BMVC, Ding_2019_ICRA, Gujjar_2019_ICRA, Scheel_2019_ICRA, Alati_2019_ICIP, Furnari_2019_ICIP, Wang_2019_ICIP, Suzuki_2018_CVPR, Casas_2018_CORL, Mahmud_2017_ICCV, Qi_2017_ICCV, Kwak_2017_IPT, Jain_2016_CVPR, Jain_2016_ICRA, Hu_2016_IROS, Schneemann_2016_IROS, Jain_2015_ICCV}, and Area Under the Curve (AUC) \cite{Rasouli_2019_BMVC, Singh_2017_ICCV, Hariyono_2016_IES} of precision-recall graph. Precision and recall also form the basis for the  calculation of some higher level metrics such as F1-score  \cite{Rasouli_2019_BMVC, Ding_2019_ICRA, Gujjar_2019_ICRA, Suzuki_2018_CVPR, Schydlo_2018_ICRA_2, Qi_2017_ICCV, Jain_2016_CVPR, Hu_2016_IROS, Schneemann_2016_IROS}, Average Precision (AP) \cite{Sun_2019_CVPR, Gujjar_2019_ICRA, Saleh_2019_ICRA, Wang_2019_ICIP, Chan_2016_ACCV} and its variations mean AP (mAP) \cite{Liang_2019_CVPR, Safaei_2019_WACV, Suzuki_2018_CVPR, Casas_2018_CORL, Zeng_2017_CVPR, Rasouli_2017_ICCVW, Gao_2017_BMVC, Singh_2017_ICCV} and calibrated AP (cAP) \cite{Gao_2017_BMVC}. Some of the less common performance metrics are Matthews Correlation Coefficient (MCC) \cite{Strickland_2018_ICRA}, False positive (FP)\cite{Jain_2015_ICCV}, True Positive Rate (TPR) and False Positive Rate (FPR) 
\cite{Volz_2015_ITSC}, Prediction Power (PP) \cite{Ziaeetabar_2018_IROS}, and Mean Reciprocal Rank (MRR) \cite{Xu_2015_ICCV}.

Depending on the application, some algorithms evaluate the timing factor in terms of the Run Time (RT) of the model \cite{Gujjar_2019_ICRA, Saleh_2019_ICRA, Volz_2016_ITSC} or time of the event, e.g. beginning of the next activity \cite{Mahmud_2017_ICCV}, Time To Accident (or collision) (TTA) \cite{Manglik_2019_IROS, Wang_2019_ICIP, Suzuki_2018_CVPR, Zeng_2017_CVPR}, and, in the context of driving, Time To Maneuver (TTM) \cite{Scheel_2019_ICRA, Wu_2019_IROS, Jain_2016_CVPR, Jain_2016_ICRA, Jain_2015_ICCV}.

\subsection{Trajectory prediction}
Perhaps the most popular performance measure for trajectory prediction is Average Displacement Error (ADE) \cite{Cho_2019_IROS, Zhou_2019_IROS, Yao_2018_CVPR, Sun_2018_ICRA, Bartoli_2018_ICPR} calculated as the average error  between the prediction location and the ground truth over all time steps. Some methods complement ADE measure with its extension Final Displacement Error (FDE) \cite{Chandra_2019_CVPR, Liang_2019_CVPR, Sadeghian_2019_CVPR, Zhao_2019_CVPR, Bi_2019_ICCV, Choi_2019_ICCV, Huang_2019_ICCV, Kosaraju_2019_NIPS, Li_2019_IROS, Chai_2019_CORL, Jain_2019_CORL, Xue_2019_WACV, Gupta_2018_CVPR, Xu_2018_CVPR_encoding, Fernando_2018_ACCV, Vemula_2018_ICRA, Xue_2018_WACV, Alahi_2016_CVPR}. As the name suggests, FDE only measures the error between the ground truth and the generated trajectory for the final time step.

Many other works use the same metric as ADE \cite{Hong_2019_CVPR, Huang_2019_ICRA, Li_2019_ICRA, Srikanth_2019_IROS, Sanchez_2019_ICIP, Bhattacharyya_2018_CVPR, Felsen_2018_ECCV, Rudenko_2018_ICRA, Rudenko_2018_IROS, Shen_2018_IROS, Solaimanpour_2017_ICRA, Yoo_2016_CVPR, Ballan_2016_ECCV, Zhou_2016_ICRA, Vasquez_2016_ICRA}  or ADE/FDE \cite{Zhang_2019_CVPR, Rasouli_2019_ICCV, Cui_2019_ICRA, Zhu_2019_IROS, Zhi_2019_CORL, Hasan_2018_CVPR, Hasan_2018_WACV, Yi_2016_ECCV} without using the same terminology. It is also a common practice that instead of using average or final time step measures, to calculate the error at different time steps over a period of time \cite{Zhao_2019_CVPR, Ding_2019_ICRA_2, Carvalho_2019_IROS, Baumann_2018_ICRA, Lee_2018_ICRA, Pfeiffer_2018_ICRA, Schulz_2018_IROS, Casas_2018_CORL, Lee_2017_CVPR, Karasev_2016_ICRA, Pfeiffer_2016_IROS, Lee_2016_WACV, Vo_2015_ICRA, Akbarzadeh_2015_IROS, Schulz_2015_ITSC}.

To measure displacement error in probablistic trajectory prediction algorithms, some works generate a set number of samples and report the best measure (i.e. minimum error)  \cite{Hong_2019_CVPR, Lee_2017_CVPR, Li_2019_CVPR, Chai_2019_CORL, Rhinehart_2019_ICCV, Rhinehart_2018_ECCV} or average over all samples \cite{Zhang_2018_CORL, Rhinehart_2019_ICCV, Rhinehart_2018_ECCV}. Depending on the error metric used, some refer to these measures as Minimum ADE/FDE (MinADE/FDE)\cite{Chang_2019_CVPR, Li_2019_CVPR, Chai_2019_CORL} (using Euclidean distance) or Mean/Minimum Mean Square Displacement (Mean/MinMSD) \cite{Rhinehart_2019_ICCV, Rhinehart_2018_ECCV} (using MSE). Some of the other probabilistic measures are Log-Likelihood (LL) \cite{Thiede_2019_ICCV, Tang_2019_ICRA, Schulz_2018_IROS}, Negative Log-Likelihood (NLL) \cite{Jain_2019_CORL, Bhattacharyya_2018_CVPR, Zhang_2018_CORL, Ma_2017_CVPR, Vasquez_2016_ICRA, Lee_2016_WACV}, Kullback–Leibler Divergence (KLD) \cite{Tang_2019_ICRA, Solaimanpour_2017_ICRA, Lee_2016_WACV}, Negative Log-Probability (NLP)\cite{Rudenko_2018_ICRA, Rudenko_2018_IROS}, Cross Entropy (CE) \cite{Rhinehart_2018_ECCV}, Average Prediction Probability (APP) \cite{Rehder_2018_ICRA}.

Performance can also be evaluated using common classification metrics. For example, in \cite{Hong_2019_CVPR, Chen_2016_ICRA} Hit Rate (HR) and in \cite{Felsen_2018_ECCV, Lee_2017_CVPR} Miss Rate (MR) metrics are used. In these cases, if the predicted trajectory is below (or above) a certain distance threshold from the groundtruth, it is considered as a hit or miss. Following a similar approach, some authors calculate accuracy \cite{Kim_2019_ICCV, Solaimanpour_2017_ICRA} or precision \cite{Yoo_2016_CVPR} of predictions.

Some of the other metrics used in the literature are Run Time (RT) \cite{Ding_2019_ICRA_2, Rudenko_2018_ICRA, Shen_2018_IROS, Vasquez_2016_ICRA, Akbarzadeh_2015_IROS}, Average Non-linear Displacement Error (ANDE)
\cite{Xu_2018_CVPR_encoding, Alahi_2016_CVPR}, Maximum Distance (MaxD) \cite{Lee_2017_CVPR, Felsen_2018_ECCV}, State collision rate (SCR) \cite{Ma_2017_CVPR}, Percentage Deviated (PD) \cite{Shkurti_2017_IROS},  Distance to Goal (DtG) \cite{Lee_2016_WACV}, Fraction of Near Misses (FNM) \cite{Bai_2015_ICRA}, Expected Calibration Error (ECE) \cite{Jain_2019_CORL}, and qualitative (Q) \cite{Mogelmose_2015_IV}. A few works predict the orientations of pedestrians \cite{Hasan_2018_CVPR, Sun_2018_ICRA, Hasan_2018_WACV} or vehicles \cite{Casas_2018_CORL}, therefore also report performance using Mean angular error (MAnE).

\subsubsection{Pitfalls of trajectory prediction metrics}
Unlike video and action prediction fields, performance measures for trajectory prediction algorithms are not standardized in terms error metrics used and units of measure. For example, for measuring displacement error, although many algorithms use Euclidean Distance (ED) (aka L2-distance, L2-norm, Euclidean norm) \cite{Liang_2019_CVPR, Sadeghian_2019_CVPR, Zhao_2019_CVPR, Bi_2019_ICCV, Kosaraju_2019_NIPS, Li_2019_IROS, Chai_2019_CORL, Jain_2019_CORL, Xue_2019_WACV, Gupta_2018_CVPR, Xu_2018_CVPR_encoding, Vemula_2018_ICRA, Xue_2018_WACV, Cho_2019_IROS, Sun_2018_ICRA, Bartoli_2018_ICPR, Srikanth_2019_IROS, Felsen_2018_ECCV, Yoo_2016_CVPR, Zhang_2019_CVPR, Cui_2019_ICRA, Zhu_2019_IROS, Zhi_2019_CORL, Hasan_2018_CVPR, Hasan_2018_WACV, Carvalho_2019_IROS, Baumann_2018_ICRA, Lee_2018_ICRA, Pfeiffer_2018_ICRA, Casas_2018_CORL, Lee_2017_CVPR, Karasev_2016_ICRA, Pfeiffer_2016_IROS, Lee_2016_WACV, Vo_2015_ICRA, Akbarzadeh_2015_IROS, Schulz_2015_ITSC, Hong_2019_CVPR, Chai_2019_CORL, Xu_2018_CVPR_encoding,  Yi_2016_ECCV, Lee_2016_WACV}, many others rely on different error metrics including MSE \cite{Huang_2019_ICCV, Alahi_2016_CVPR, Zhou_2019_IROS, Yao_2018_CVPR, Huang_2019_ICRA, Sanchez_2019_ICIP, Bhattacharyya_2018_CVPR, Solaimanpour_2017_ICRA, Rasouli_2019_ICCV, Rhinehart_2019_ICCV, Rhinehart_2018_ECCV}, RMSE \cite{Chandra_2019_CVPR, Hong_2019_CVPR, Zhao_2019_CVPR, Huang_2019_ICRA, Zhao_2019_CVPR, Ding_2019_ICRA_2}, Weighted RMSE \cite{Schulz_2018_IROS}, Mean Absolute Error (MAE)\cite{Casas_2018_CORL, Li_2019_ICRA}, Hausdorff Distance (HD)\cite{Zhang_2018_CORL}, Modified HD (MHD) \cite{Rudenko_2018_ICRA, Rudenko_2018_IROS, Shen_2018_IROS, Yoo_2016_CVPR, Ballan_2016_ECCV, Vasquez_2016_ICRA, Lee_2016_WACV}, and discrete Fr\'echet distance (DFD) \cite{Zhi_2019_CORL}. Moreover, trajectory prediction algorithms use different units for measuring displacement error. These are meter \cite{Chandra_2019_CVPR, Liang_2019_CVPR, Sadeghian_2019_CVPR, Zhao_2019_CVPR, Kosaraju_2019_NIPS, Chai_2019_CORL, Jain_2019_CORL, Gupta_2018_CVPR, Vemula_2018_ICRA, Cho_2019_IROS, Hong_2019_CVPR, Huang_2019_ICRA, Li_2019_ICRA, Srikanth_2019_IROS, Sanchez_2019_ICIP, Shen_2018_IROS, Zhang_2019_CVPR, Cui_2019_ICRA, Zhi_2019_CORL, Hasan_2018_CVPR, Hasan_2018_WACV, Zhao_2019_CVPR, Ding_2019_ICRA_2, Carvalho_2019_IROS, Baumann_2018_ICRA, Lee_2018_ICRA, Pfeiffer_2018_ICRA, Casas_2018_CORL, Lee_2017_CVPR, Karasev_2016_ICRA, Pfeiffer_2016_IROS, Akbarzadeh_2015_IROS, Schulz_2015_ITSC, Hong_2019_CVPR, Lee_2017_CVPR, Chai_2019_CORL, Rhinehart_2018_ECCV}, pixel \cite{Sadeghian_2019_CVPR, Zhao_2019_CVPR, Bi_2019_ICCV, Choi_2019_ICCV, Li_2019_IROS, Xue_2019_WACV, Bhattacharyya_2018_CVPR, Ballan_2016_ECCV, Lee_2017_CVPR}, normalized pixel \cite{Choi_2019_ICCV, Xue_2018_WACV, Lee_2016_WACV} and feet \cite{Felsen_2018_ECCV}.

Although such discrepancy between error metrics and units is expected across different applications, the problem arises when the proposed works do not specify the error metric \cite{Choi_2019_ICCV, Li_2019_CVPR}, the unit of measure \cite{Huang_2019_ICCV, Xu_2018_CVPR_encoding,  Alahi_2016_CVPR, Zhou_2019_IROS, Yao_2018_CVPR, Rudenko_2018_ICRA, Rudenko_2018_IROS, Solaimanpour_2017_ICRA, Yoo_2016_CVPR, Vasquez_2016_ICRA, Zhu_2019_IROS, Yi_2016_ECCV, Schulz_2018_IROS, Zhang_2018_CORL, Li_2019_CVPR, Xu_2018_CVPR_encoding} or both \cite{Fernando_2018_ACCV, Zhou_2016_ICRA}. Despite the fact that the reported results might imply the choice of the metrics and units, the lack of specification can cause erroneous comparisons, specially because many authors use the results of previous works directly as reported in the papers.

Unfortunately, metric and unit discrepancy exists within the same applications and the same error measuring techniques. For instance, in the case of ADE measure, this metric is originally proposed in \cite{Pellegrini_2009_ICCV} in terms of ED, and was referred to as ADE by the authors of \cite{Alahi_2016_CVPR} despite the fact that they used MSE instead. This is also apparent in many subsequent works that employed ADE measure. For example, the majority of methods use the original metric and report the results in terms of ED \cite{Liang_2019_CVPR, Sadeghian_2019_CVPR, Bi_2019_ICCV, Xue_2019_WACV, Gupta_2018_CVPR, Xu_2018_CVPR_encoding, Kosaraju_2019_NIPS, Li_2019_IROS, Chai_2019_CORL, Jain_2019_CORL, Vemula_2018_ICRA, Xue_2018_WACV, Cho_2019_IROS, Sun_2018_ICRA, Bartoli_2018_ICPR}  whereas some works use MSE \cite{Huang_2019_ICCV, Alahi_2016_CVPR, Zhou_2019_IROS, Yao_2018_CVPR} and RMSE\cite{Chandra_2019_CVPR, Zhao_2019_CVPR} or do not specify the metric  \cite{Choi_2019_ICCV, Fernando_2018_ACCV}. Although the formulation of these metrics look similar, they produce different results. ADE using ED, for example, is square-root of squared differences averaged over all samples and time steps. Unlike ED, in RMSE, the averaging takes place inside square-root operation. MSE, on the other hand, is very different from the other two metrics, and does not calculate the root of the error. As we can also see in some of the past works, this discrepancy may cause confusion about the intended and actual metric that is used. For example, in \cite{Zhao_2019_CVPR} the authors propose to use MAE metric while presenting mathematical formulation of Euclidean distance. The authors of \cite{Bartoli_2018_ICPR, Yi_2016_ECCV} make a similar mistake and define ED formulation but refer to it as MSE.

In addition, some algorithms within the same applications and using the same datasets use different measuring unit. For instance, in the context of surveillance, ETH \cite{Pellegrini_2009_ICCV} is one of the most commonly used datasets. Many works use this dataset for benchmarking the performance of their proposed algorithms, however, they either use different units, e.g. meter \cite{Liang_2019_CVPR, Sadeghian_2019_CVPR, Zhang_2019_CVPR, Kosaraju_2019_NIPS, Anderson_2019_IROS, Gupta_2018_CVPR, Pfeiffer_2018_ICRA, Vemula_2018_ICRA}, pixel \cite{Li_2019_IROS, Zhao_2019_CVPR, Choi_2019_ICCV, Xue_2019_WACV}, normalized pixel \cite{Xue_2018_WACV, Choi_2019_ICCV}, or do not specify the unit used \cite{Huang_2019_ICCV, Li_2019_CVPR, Zhu_2019_IROS, Xu_2018_CVPR_encoding, Fernando_2018_ACCV, Alahi_2016_CVPR}.

Last but not least, another potential source of error in performance evaluation is in the design of the experiments. Taking surveillance applications as an example, it is a common practice to evaluate algorithms with 8 frames observations of the past and prediction 12 steps in the future  \cite{Li_2019_CVPR, Liang_2019_CVPR, Sadeghian_2019_CVPR, Zhang_2019_CVPR, Gupta_2018_CVPR, Vemula_2018_ICRA, Alahi_2016_CVPR}. However, in some cases the performance of state-of-the-art is reported under the standard 8/12 condition, but the proposed algorithms are tested under different conditions. For instance, in \cite{Xu_2018_CVPR_encoding} the authors incorrectly compared the performance of their proposed algorithm using 5 observations and 5 predictions with the results of the previous works evaluated under the standard 8/12 condition.

\subsection{Motion prediction}
Due to the inherent stochasticity of human body movement, motion prediction algorithms often limit their prediction horizon to approximately $500ms$. To measure the error between corresponding ground truth and predicted poses, these algorithms use mean average error, either in angle space (MAnE) \cite{Gopalakrishnan_2019_CVPR, Liu_2019_CVPR, Mao_2019_ICCV, Hernandez_2019_ICCV, Wang_2019_ICCV, Chiu_2019_WACV, Gui_2018_ECCV, Gui_2018_ECCV_2, Gui_2018_IROS, Martinez_2017_CVPR, Jain_2016_CVPR, Fragkiadaki_2015_ICCV} or joint space (MJE) in terms of joint coordinates \cite{Liu_2019_CVPR, Wu_2019_WACV, Yao_2018_CVPR, Butepage_2018_ICRA, Joo_2019_CVPR, Butepage_2018_ICRA, Walker_2017_ICCV, Butepage_2017_CVPR, Chiu_2019_WACV, Yuan_2019_ICCV, Talignani_2019_IROS}. In the 3D motion prediction domain, a metric known as Mean Per Joint Prediction Error (MPJPE) \cite{Mao_2019_ICCV, Zhang_2019_ICCV} is used which is the error over joints normalized with respect to the root joint. 

As an alternative to distance error metrics, Percentage of Correct Keypoints (PCK) \cite{Zhang_2019_ICCV, Chiu_2019_WACV, Chao_2017_CVPR, Wu_2019_WACV} measures how many of the keypoints (e.g. joints) are predicted correctly. The correct predictions are those that are below a certain error threshold (e.g. 0.05). Some works also use the accuracy metric to report on how well the algorithm can localize the position of a particular joint within an error tolerance region \cite{Luo_2019_IROS, Fragkiadaki_2015_ICCV}.

Other metrics used in the literature include Normalized Power Spectrum Similarity (NPSS) \cite{Gopalakrishnan_2019_CVPR}, Reconstruction Error (RE) \cite{Zhang_2019_ICCV}, Limb Orientation (LO) \cite{Joo_2019_CVPR}, PoSe Entropy (PSEnt), PoSe KL (PSKL) \cite{Hernandez_2019_ICCV}, qualitative human judgment \cite{Hernandez_2019_ICCV, Gui_2018_ECCV_2} and method Run Time (RT) \cite{Liu_2019_CVPR, Wu_2019_WACV}.

\subsubsection{Pitfalls of motion prediction metrics}
Similar to trajectory methods, motion prediction algorithms are evaluated using distance-based methods that calculate the error between pose vectors. In the case of MAnE measure, some methods use ED metric \cite{Mao_2019_ICCV, Wang_2019_ICCV, Gui_2018_IROS, Martinez_2017_CVPR, Jain_2016_CVPR, Fragkiadaki_2015_ICCV} while others use MSE \cite{Gopalakrishnan_2019_CVPR, Liu_2019_CVPR, Hernandez_2019_ICCV, Gui_2018_ECCV, Gui_2018_ECCV_2}. Sometimes no metric is specified  \cite{Chiu_2019_WACV}. The same holds for MJE measure where metrics used include MSE \cite{Liu_2019_CVPR, Yao_2018_CVPR, Butepage_2018_ICRA, Butepage_2018_ICRA}, RMSE \cite{Wu_2019_WACV, Talignani_2019_IROS}, ED \cite{Walker_2017_ICCV, Butepage_2017_CVPR, Yuan_2019_ICCV}, MAE \cite{Chiu_2019_WACV}, or no metric is specified \cite{Joo_2019_CVPR}.

The added challenge in coordinate-based error measures, e.g. MJE, MPJPE, is the error unit. While many approaches do not specify the unit explicitly \cite{Liu_2019_CVPR, Butepage_2018_ICRA, Butepage_2018_ICRA, Walker_2017_ICCV, Butepage_2017_CVPR, Chiu_2019_WACV, Yuan_2019_ICCV}, others clearly state whether the unit is in pixel \cite{Wu_2019_WACV, Yao_2018_CVPR}, centimeter \cite{Joo_2019_CVPR}, meter \cite{Talignani_2019_IROS} or millimeter \cite{Mao_2019_ICCV, Zhang_2019_ICCV}. As was the case before, here many algorithms that benchmark on the same datasets, may use different performance metrics, e.g. using popular human datasset  Human 3.6M \cite{Ionescu_2014_PAMI}, MAnE (ED), MAnE (MSE)\cite{Gui_2018_ECCV} and MJE (ED) \cite{Gui_2018_IROS} are used.

\subsection{Other prediction applications}
Depending on the task objectives, the metrics used in other prediction applications are similar to the ones discussed thus far. For instance, the applications that classify future events or outcomes, e.g. contest or an election winner, next image index for storytelling, severe weather, and pain, use common metrics such as accuracy  \cite{Carvajal_2016_ICPR, Joo_2015_ICCV, Zeng_2017_ICCV}, precision, recall \cite{Kim_2019_WACV, Sur_2017_IROS}, percentage of correct predictions (PCP) \cite{Mottaghi_2016_ECCV}, and Matthews correlation coefficient (MCC) \cite{Sur_2017_IROS} which predicts the quality of binary classification by taking into account both false and true predictions.

Regression based methods, such as temperature, trends, or steering prediction, use distance metrics including Euclidean Distance (ED) \cite{Graves_2019_IROS, Zapf_2019_IROS}, RMSE \cite{Chu_2018_WACV}, MSE \cite{Jin_2017_NIPS}, MAE \cite{He_2018_ICPR, Al-Halah_2017_ICCV, Wilson_2015_ICRA}, Mean Absolute Percentage Error (MAPE) \cite{Wang_2018_WACV,Al-Halah_2017_ICCV}, normalized MAPE (nMAPE) \cite{Siddiqui_2019_WACV}, and the Spearman's ranking Correlation (SRC) \cite{Wang_2018_WACV}  which measures the strength and direction of relationship between two variables.

Of particular interest are metrics used for evaluating generative models that predict Occupancy Grid Maps (OGMs) and segmentation maps. OGMs are grayscale images that highlight the likelihood of a certain region (represented as a cell in the grid) that is occupied. The generated map can be compared to ground truth by using image similarity metrics such as SSIM \cite{Mohajerin_2019_CVPR, Katyal_2019_ICRA}, PSNR \cite{Katyal_2019_ICRA} or psi ($\psi$)  \cite{Afolabi_2018_IROS}. Alternatively, OGM can be evaluated using a binary classification metric. Here, the grid cells are classified as occupied or free by applying a threshold and then can be evaluated as a whole by using metrics such as True Positive (TP), True Negative (TN) \cite{Mohajerin_2019_CVPR}, Receiver Operator Characteristic (ROC) curve over TP and TN \cite{Schreiber_2019_ICRA, Hoermann_2018_ICRA}, F1-score \cite{Guizilini_2019_ICRA, Schreiber_2019_ICRA}, precision, recall, and their corresponding AUC \cite{Choi_2016_IROS}. Given that OGM prediction algorithms are mainly used in safety-critical applications such as autonomous driving, some algorithms  are also evaluated in terms of their Run Time (RT) \cite{Mohajerin_2019_CVPR, Katyal_2019_ICRA, Choi_2016_IROS}.

Image similarity metrics such as PSNR and SSIM can also be used in the segmentation prediction domain \cite{Luc_2017_ICCV}. The most common metric, however, is Intersection over Union (IoU)\cite{Terwilliger_2019_WACV, Luc_2018_ECCV, Luc_2017_ICCV, Jin_2017_NIPS} which measures the average overlap of segmented instances with the ground truth segments. In addition, by applying a threshold to IoU scores, the true matches can be identified and used to calculate the Average Precision (AP) scores as in \cite{Luc_2018_ECCV}. Other metrics used for segmentation prediction tasks include EndPoint error (EPE) \cite{Jin_2017_NIPS}, Probabilistic Rand Index (RI), Global Consistency Error (GCE), and Variation of Information (VoI) \cite{Luc_2018_ECCV}.

\section{Datasets}
\label{datasets}

\begin{table*}[]
\centering
\resizebox{1\textwidth}{!}{%
\begin{tabular}{|l|l|l|l|c|c|c|c|c|}
\hline
\multirow{2}{*}{\textbf{Year}} & \multicolumn{1}{c|}{\multirow{2}{*}{\textbf{Dataset}}} & \multicolumn{1}{c|}{\multirow{2}{*}{\textbf{Type}}} & \multicolumn{1}{c|}{\multirow{2}{*}{\textbf{Annotations}}} & \multicolumn{5}{c|}{\textbf{Application}} \\ \cline{5-9} 
 & \multicolumn{1}{c|}{} & \multicolumn{1}{c|}{} & \multicolumn{1}{c|}{} & \textit{\textbf{V}} & \textit{\textbf{A}} & \textit{\textbf{T}} & \textit{\textbf{M}} & \textit{\textbf{O}} \\ \hline
\multirow{12}{*}{\textit{\textbf{2019}}} & ARGOVerse \cite{Chang_2019_CVPR} & Traffic & RGB, LIDAR, 3D BB &  &  & x &  &  \\ \cline{2-9} 
 & CARLA \cite{Rhinehart_2019_ICCV} & Traffic (sim) & RGB &  & x &  &  &  \\ \cline{2-9} 
 & EgoPose \cite{Yuan_2019_ICCV} & Pose (ego) & RGB, 3D Pose &  &  &  & x &  \\ \cline{2-9} 
 & Future Motion (FM) \cite{Kim_2019_ICCV} & Mix & RGB, BB, Attrib. &  &  & x &  &  \\ \cline{2-9} 
 & InstaVariety \cite{Kanazawa_2019_CVPR} & Activities & RGB, BB, Pose &  &  &  & x &  \\ \cline{2-9} 
 & INTEARCTION \cite{Zhan_2019_arxiv} & Traffic & Map, Traj. &  &  & x &  &  \\ \cline{2-9} 
 & Luggage \cite{Manglik_2019_IROS} & Robot & Stereo RGB, BB &  & x &  &  &  \\ \cline{2-9} 
 & MGIF \cite{Siarohin_2019_CVPR} & Activities & RGB & x &  &  &  &  \\ \cline{2-9} 
 & Pedestrian Intention Estimation (PIE) \cite{Rasouli_2019_ICCV} & Traffic & RGB, BB, Class, Attrib., Temporal seg., Vehicle sensors &  & x & x &  &  \\ \cline{2-9} 
 & nuScenes \cite{Caesar_2019_arxiv} & Traffic & RGB, LIDAR, 3D BB, Vehicle sensors &  &  & x &  &  \\ \cline{2-9} 
 & Vehicle-Pedestrian-Mixed (VPM) \cite{Bi_2019_ICCV} & Traffic & RGB, BB &  &  & x &  &  \\ \cline{2-9} 
 & TRAF \cite{Chandra_2019_CVPR} & Traffic & RGB, BN, Class, Time-of-day &  &  & x &  &  \\ \hline
\multirow{10}{*}{\textit{\textbf{2018}}} & 3D POSES IN THE WILD (3DPW) \cite{vonMarcard_2018_ECCV} & Outdoor & RGB, 2D/3D Pose, Models &  &  &  & x &  \\ \cline{2-9} 
 & ActEV/VIRAT \cite{Awad_2018_Trecvid} & Surveillance & RGB, BB, Activity, Temporal seg. &  & x & x &  &  \\ \cline{2-9} 
 & ACTICIPATE \cite{Schydlo_2018_ICRA} & Interaction & RGB, Gaze, Pose &  & x &  &  &  \\ \cline{2-9} 
 & Atomic Visual Actions (AVA) \cite{Gu_2018_CVPR} & Activities & RGB, Activity, Temporal seg. &  & x &  &  &  \\ \cline{2-9} 
 & Epic-Kitchen \cite{Damen_2018_ECCV} & Cooking  (ego) & RGB, Audio,  BB, Class, Text, Temporal seg. &  & x &  &  &  \\ \cline{2-9} 
 & EGTEA Gaze+ \cite{Li_2018_ECCV_2} & Cooking  (ego) & RGB, Gaze, Mask, Activity, Temporal seg. &  & x &  &  &  \\ \cline{2-9} 
 & ShanghaiTech Campus (STC) \cite{Liu_2018_CVPR} & Surveillance & RGB, Anomaly & x &  &  &  &  \\ \cline{2-9} 
 & ShapeStack \cite{Groth_2018_arxiv} & Objects (sim) & RGBD, Mask, Stability & x &  &  &  &  \\ \cline{2-9} 
 & VIENA \cite{Aliakbarian_2018_ACCV} & Traffic (sim) & RGB, Activity, Vehicle sensors &  & x &  &  &  \\ \cline{2-9} 
 & YouCook2 \cite{Zhou_2018_AI} & Cooking & RGB, Audio, Text, Activity, Temporal seg. &  & x &  &  &  \\ \hline
\multirow{10}{*}{\textit{\textbf{2017}}} & BU Action (BUA) \cite{Ma_2017_PR} & Activities & RGB (image), Activity &  & x &  &  &  \\ \cline{2-9} 
 & CityPerson \cite{Shanshan_2017_CVPR} & Traffic & Stereo RGB, BB, Semantic seg. &  &  & x &  &  \\ \cline{2-9} 
 & Epic-Fail \cite{Zeng_2017_CVPR} & Risk assessment & RGB, BB, Traj., Temporal seg. &  & x &  &  &  \\ \cline{2-9} 
 & Joint Attention in Autonomous Driving (JAAD) \cite{Rasouli_2017_ICCVW} & Traffic & RGB, BB, Attrib., Temporal seg. & x & x & x &  & x \\ \cline{2-9} 
 & L-CAS \cite{Yan_2017_IROS} & Traffic & LIDAR, 3D BB, Attrib. &  &  & x &  &  \\ \cline{2-9} 
 & Mouse Fish \cite{Xu_2017_IJCV} & Animals & Depth, 3D Pose &  &  &  & x &  \\ \cline{2-9} 
 & Oxford Robot Car (ORC) \cite{Maddern_2017_IJRR} & Traffic & Stereo RGB, LIDAR, Vehicle sensors &  &  & x &  &  \\ \cline{2-9} 
 & PKU-MMD \cite{Liu_2017_arxiv} & Activities, interactions & RGBD, IR, 3D Pose, Multiview, Temporal seg. &  & x &  &  &  \\ \cline{2-9} 
 & Recipe1M \cite{Salvador_2017_CVPR} & Cooking & RGB(image), Text &  & x &  &  &  \\ \cline{2-9} 
 & STRANDS \cite{Hawes_2017_RAM} & Traffic & RGBD, 3DBB &  &  & x &  &  \\ \hline
\multirow{13}{*}{\textit{\textbf{2016}}} & BAIR Push \cite{Finn_2016_NIPS} & Object manipulation & RGB & x &  &  &  &  \\ \cline{2-9} 
 & Bouncing Ball (BB) \cite{Chang_2016_arxiv} & Simulation & RGB & x &  &  &  &  \\ \cline{2-9} 
 & Miss Universe (MU) \cite{Carvajal_2016_ICPR} & Miss universe & RGB, BB, Scores &  &  &  &  & x \\ \cline{2-9} 
 & Cityscapes \cite{Cordts_2016_CVPR} & Traffic & Stereo RGB, BB, Semantic seg., Vehicle Sensors & x &  & x &  & x \\ \cline{2-9} 
 & CMU Mocap \cite{CMU_Mocap_2016} & Activities & 3D Pose, Activity &  & x &  & x &  \\ \cline{2-9} 
 & Dashcam Accident Dataset (DAD) \cite{Chan_2016_ACCV} & Traffic, accidents & RGB, BB, Class,  Temporal seg. &  & x &  &  &  \\ \cline{2-9} 
 & NTU RGB-D \cite{Shahroudy_2016_CVPR} & Activities & RGBD, IR, 3D Pose, Activity &  & x &  &  &  \\ \cline{2-9} 
 & Ongoing Activity (OA) \cite{Li_2016_WACV} & Actvities & RGB, Activity &  & x &  &  &  \\ \cline{2-9} 
 & OAD \cite{Li_2016_ECCV} & Activities & RGBD, 3D Pose, Activity, Temporal seg. &  & x &  &  &  \\ \cline{2-9} 
 & Stanford Drone (SD) \cite{Robicquet_2016_ECCV} & Surveillance & RGB, BB, Class &  &  & x &  &  \\ \cline{2-9} 
 & TV Series \cite{De_2016_ECCV} & Activities & RGB, Activity, Temporal seg. &  & x &  &  &  \\ \cline{2-9} 
 & Visual StoryTelling (VIST) \cite{Huang_2016_NAACL} & Visual story & RGB, Text &  &  &  &  & x \\ \cline{2-9} 
 & Youtube-8M \cite{Abu_2016_arxiv} & Activities & RGB, Activity, Temporal seg. & x &  &  &  &  \\ \hline
\multirow{14}{*}{\textit{\textbf{2015}}} & Amazon \cite{Mcauley_2015_CRDIR} & Fashion & Features, Attrib., Text &  &  &  &  & x \\ \cline{2-9} 
 & Atari \cite{Oh_2015_NIPS} & Games & RGB & x &  &  &  &  \\ \cline{2-9} 
 & Brain4Cars \cite{Jain_2015_ICCV} & Traffic, Driver & RGB, BB, Attrib., Temporal seg., Vehicle sensors &  & x &  &  &  \\ \cline{2-9} 
 & CMU Panoptic \cite{Joo_2015_ICCV_2} & Interaction & RGBD, Multiview, 3D Pose, 3D facial landmark & x & x &  & x &  \\ \cline{2-9} 
 & First Person Personalized Activities (FPPA) \cite{Zhou_2015_ICCV} & Activities (ego) & RGB, Activity, Temporal seg. &  & x &  &  &  \\ \cline{2-9} 
 & GTEA Gaze + \cite{Li_2015_CVPR} & Cooking (ego) & RGB, Gaze, Mask, Activity, Temporal seg. &  & x &  &  &  \\ \cline{2-9} 
 & MicroBlog-Images (MBI-1M) \cite{Cappallo_2015_ICMR} & Tweets & RGB (image), Attrib., Text &  &  &  &  & x \\ \cline{2-9} 
 & MOT \cite{Leal_2015_arxiv} & Surveillance & RGB, BB &  &  & x &  &  \\ \cline{2-9} 
 & Moving MNIST (MMNIST) \cite{Srivastava_2015_ICML} & Digits & Grayscale & x &  &  &  &  \\ \cline{2-9} 
 & SUN RGB-D \cite{Song_2015_CVPR} & Places & RGBD, 3D BB , Class &  &  &  &  & x \\ \cline{2-9} 
 & SYSU 3DHOI \cite{Hu_2015_CVPR} & Object interaction & RGBD, 3D Pose, Activity &  & x &  &  &  \\ \cline{2-9} 
 & THUMOS \cite{Gorban_2015} & Activities & RGB, Activity, Temporal seg. & x & x &  &  &  \\ \cline{2-9} 
 & Watch-n-Push (WnP) \cite{Wu_2015_CVPR} & Activities & RGBD, 3D Pose, Activity, Temporal seg. &  & x &  &  &  \\ \cline{2-9} 
 & Wider \cite{Xiong_2015_CVPR} & Activities & RGB (image), Activity &  & x &  &  &  \\ \hline
\multirow{5}{*}{\textit{\textbf{2014}}} & Breakfast \cite{Kuehne_2014_CVPR} & Cooking & RGB, Activity, Temporal seg. &  & x &  &  &  \\ \cline{2-9} 
 & Human3.6M \cite{Ionescu_2014_PAMI} & Activities & RGB, 3D Pose, Activity & x & x &  & x &  \\ \cline{2-9} 
 & MPII Human Pose \cite{Andriluka_2014_CVPR} & Activities & RGB, Pose, Activity &  &  &  & x &  \\ \cline{2-9} 
 & Online RGBD Action Dataset (ORGBD) \cite{Yu_2014_ACCV} & Activities & RGBD, BB, 3D Pose, Activity &  & x &  &  &  \\ \cline{2-9} 
 & Sports-1M \cite{Karpathy_2014_CVPR} & Sports & RGB, Activity & x & x &  &  &  \\ \hline

\end{tabular}
}
\caption{A summary of common datasets from years 2014-2019 used in vision-based prediction applications, namely video (V), action (A), trajectory (T), motion (M) and others (O). The annotation column specifies the type of data (e.g. RGB, Infrared(IR)) and annotation types. All datasets contain image sequences unless specified by "image". As for annotations, BB stands for bounding box. Attributes include any object characteristics (e.g. for pedestrians demographics, behavior). Vehicle sensors may include speed, steering angle, GPS, etc. Temporal seg. identifies the datasets that specify the start and end of the events.}
\label{dataset_table1}
\end{table*}

\begin{table*}[]
\centering
\resizebox{1\textwidth}{!}{%
\begin{tabular}{|l|l|l|l|c|c|c|c|c|}
\hline
\multirow{2}{*}{\textbf{Year}} & \multicolumn{1}{c|}{\multirow{2}{*}{\textbf{Dataset}}} & \multicolumn{1}{c|}{\multirow{2}{*}{\textbf{Type}}} & \multicolumn{1}{c|}{\multirow{2}{*}{\textbf{Annotations}}} & \multicolumn{5}{c|}{\textbf{Application}} \\ \cline{5-9} 
 & \multicolumn{1}{c|}{} & \multicolumn{1}{c|}{} & \multicolumn{1}{c|}{} & \textit{\textbf{V}} & \textit{\textbf{A}} & \textit{\textbf{T}} & \textit{\textbf{M}} & \textit{\textbf{O}} \\ \hline
 \multirow{7}{*}{\textit{\textbf{2013}}} & 50Salads \cite{Stein_2013_IJCPUC} & Cooking  (ego) & RGBD, Activity, Temporal seg. &  & x &  &  &  \\ \cline{2-9} 
 & ATC \cite{Brvsvcic_2013_HMS} & Surveillance & RGB, Traj., Attrib. &  &  & x &  &  \\ \cline{2-9} 
 & CAD-120 \cite{Koppula_2013_IJRR} & Activities & RGBD, 3D Pose, Activity &  & x &  &  &  \\ \cline{2-9} 
 & CHUK Avenue \cite{Lu_2013_ICCV} & Surveillance & RGB, BB, Anomaly, Temporal seg. & x &  & x &  &  \\ \cline{2-9} 
 & Daimler Path \cite{Schneider_2013_GCPR} & Traffic & Stereo Grayscale, BB, Temporal seg. , Vehicle sensors &  & x &  &  &  \\ \cline{2-9} 
 & Joint-annotated HMDB (JHMDB) \cite{Jhuang_2013_ICCV} & Activities & RGB, Mask, Activity, Pose, Optical flow & x & x &  &  &  \\ \cline{2-9} 
 & Penn Action \cite{Zhang_2013_ICCV} & Activities & RGB, BB, Pose, Activity & x &  &  & x &  \\ \hline
\multirow{11}{*}{\textit{\textbf{2012}}} & BIT \cite{Kong_2012_ECCV} & Interaction & RGB, Activity &  & x &  &  &  \\ \cline{2-9} 
 & GTEA Gaze \cite{Fathi_2012_ECCV} & Cooking  (ego) & RGB, Gaze,  Mask, Activity, Temporal seg. &  & x &  &  &  \\ \cline{2-9} 
 & KITTI \cite{Geiger_2012_CVPR} & Traffic & Stereo RGB, LIDAR, BB,  Optical flow, Vehicle sensors & x &  & x &  & x \\ \cline{2-9} 
 & MANIAC \cite{Abramov_2012_WACV} & Object manipulation & RGBD, Semantic seg., Activity &  & x &  &  &  \\ \cline{2-9} 
 & MPII-Cooking \cite{Rohrbach_2012_CVPR} & Cooking & RGB, 3D Pose, Activity, Temporal seg. &  & x &  &  &  \\ \cline{2-9} 
 & MSR Daily Activity (MSRDA) \cite{Wang_2012_CVPR} & Activities & Depth, Activity &  & x &  &  &  \\ \cline{2-9} 
 & New York Grand Central (GC) \cite{Zhou_2012_CVPR} & Surveillance & RGB, Traj. &  &  & x &  &  \\ \cline{2-9} 
 & SBU Kinetic Interction (SBUKI) \cite{kiwon_2012_CVPR} & Interaction & RGBD, 3D Pose, Activity &  & x & x & x &  \\ \cline{2-9} 
 & UCF-101 \cite{Soomro_2012_arxiv} & Activities & RGB, Activity & x & x &  & x &  \\ \cline{2-9} 
 & UTKinect-Action (UTKA) \cite{Xia_2012_CVPRW} & Activities & RGBD, 3D Pose, Activity, Temporal seg. &  & x &  &  &  \\ \cline{2-9} 
 & UvA-NEMO \cite{Dibeklio_2012_ECCV} & Smiles & RGB & x &  &  &  &  \\ \hline
\multirow{5}{*}{\textit{\textbf{2011}}} & Ford campus vision LiDAR (FCVL) \cite{Pandey_2011_IJRR} & Traffic & RGB, LIDAR, Vehicle sensors &  &  &  &  & x \\ \cline{2-9} 
 & Human Motion Database (HMDB) \cite{Kuehne_2011_ICCV} & Activities & RGB, BB, Mask, Activity &  & x &  &  &  \\ \cline{2-9} 
 & Stanford 40 \cite{Yao_2011_ICCV} & Activities & RGB (image), BB, Activity &  & x &  &  &  \\ \cline{2-9} 
 & Town Center \cite{Benfold_2011_CVPR} & Surveillance & RGB, BB &  &  & x &  &  \\ \cline{2-9} 
 & VIRAT \cite{Oh_2011_CVPR} & Surveillance, Activities & RGB, BB, Activity, Temporal seg. &  & x & x &  &  \\ \hline
\multirow{8}{*}{\textit{\textbf{2010}}} & DISPLECS \cite{Pugeault_2010_ECCV} & Traffic & RGB, Vehicle sensors &  &  &  &  & x \\ \cline{2-9} 
 & MSR \cite{Li_2010_CVPRW} & Activities & Depth, Activity & x &  &  &  &  \\ \cline{2-9} 
 & MUG \cite{Aifanti_2010_WIAMIS} & Facial expressions & RGB, Keypoints, Label & \multicolumn{1}{l|}{x} & \multicolumn{1}{l|}{} & \multicolumn{1}{l|}{} & \multicolumn{1}{l|}{} & \multicolumn{1}{l|}{} \\ \cline{2-9} 
 & PROST \cite{Santner_2010_CVPR} & Objects & RGB, BB & x &  &  &  &  \\ \cline{2-9} 
 & TV Human Interaction (THI) \cite{Patron_2010_BMVC} & Interaction & RGB, BB, Head pose, Activity &  & x &  &  &  \\ \cline{2-9} 
 & UT Interaction (UTI) \cite{Ryoo_2010_UT} & Interaction & RGB, BB, Activity, Temporal seg. &  & x &  &  &  \\ \cline{2-9} 
 & VISOR \cite{Vezzani_2010_MTA} & Surveillance & RGB, BB, Pose, Attrib. & x &  &  &  &  \\ \cline{2-9} 
 & Willow Action \cite{Delaitre_2010_BMVC} & Activiites & RGB (image), Activity &  & x &  &  &  \\ \hline
\multirow{9}{*}{\textit{\textbf{2009}}} & Caltech Pedestrian \cite{Dollar_2009_CVPR} & Traffic & RGB, BB & x & x &  &  &  \\ \cline{2-9} 
 & Collective Activity (CA) \cite{Choi_2009_ICCVW} & Interaction & RGB, BB, Attrib., Activity, Temporal seg. &  & x & x & x &  \\ \cline{2-9} 
 & Edinburgh IFP \cite{Majecka_2009} & Surveillance & RGB, BB &  &  & x &  &  \\ \cline{2-9} 
 & ETH \cite{Pellegrini_2009_ICCV} & Surveillance & RGB, Traj. &  &  & x &  &  \\ \cline{2-9} 
 & OSU \cite{Hess_2009_CVPR} & Sports & RGB, BB, Attrib. &  &  & x &  &  \\ \cline{2-9} 
 & PETS2009 \cite{Ferryman_2009_PETS} & Surveillance & RGB, BB &  &  & x &  &  \\ \cline{2-9} 
 & QMUL \cite{Loy_2009_BMVC} & Traffic, anomaly & RGB, Traj. &  &  & x &  &  \\ \cline{2-9} 
 & TUM Kitchen \cite{Tenorth_2009_ICCVW} & Activities & RGB, RFID, 3D Pose, Activity, Temporal seg. &  &  & x &  &  \\ \cline{2-9} 
 & YUV Videos \cite{ASU_2009_YUV} & Mix videos & RGB & x &  &  &  &  \\ \hline
\multirow{2}{*}{\textit{\textbf{2008}}} & Daimler \cite{Enzweiler_2008_PAMI} & Traffic & Grayscale, BB &  & x &  &  &  \\ \cline{2-9} 
 & MIT Trajectory (MITT) \cite{Grimson_2008_CVPR} & Surveillance & RGB, Traj. &  &  & x &  &  \\ \hline
\multirow{5}{*}{\textit{\textbf{2007}}} & AMOS \cite{Jacobs_2007_CVPR} & Weather & RGB, Temperature, Time &  &  &  &  & x \\ \cline{2-9} 
 & ETH Pedestrian \cite{Ess_2007_ICCV} & Traffic & RGB, BB &  & x &  &  &  \\ \cline{2-9} 
 & Lankershim Boulevard \cite{US_2007_Lankershim} & Traffic & RGB, Traj. &  &  & x &  &  \\ \cline{2-9} 
 & Next Generation Simulation (NGSIM) \cite{NGSIM_2007} & Traffic & Map, Traj. &  & x & x &  &  \\ \cline{2-9} 
 & UCY \cite{Lerner_2007_CGF} & Surveillance & RGB, Traj., Gaze &  &  & x &  &  \\ \hline
\textit{\textbf{2006}} & Tuscan Arizona \cite{Pickering_2006} & Weather & RGB &  &  &  &  & x \\ \hline
\textit{\textbf{2004}} & KTH \cite{Schuldt_2004_ICPR} & Activities & Grayscale, Activity & x &  &  &  &  \\ \hline
\textit{\textbf{1981}} & Golden Colorado \cite{Stoffel_1981} & Weather & RGB &  &  &  &  & x \\ \hline
\end{tabular}
}
\caption{A summary of common datasets from years 2013 and earlier used in vision-based prediction applications, namely video (V), action (A), trajectory (T), motion (M) and others (O). The annotation column specifies the type of data (e.g. RGB, Infrared(IR)) and annotation types. All datasets contain sequences unless specified by "image". As for annotations, BB stands for bounding box. Attributes include any object characteristics (e.g. for pedestrians demographics, behavior). Vehicle sensors may include speed, steering angle, GPS, etc. Temporal seg. identifies the datasets that specify the start and end of the events.}
\label{dataset_table2}
\end{table*}
\begin{figure*}
\centering
\includegraphics[width=1\textwidth]{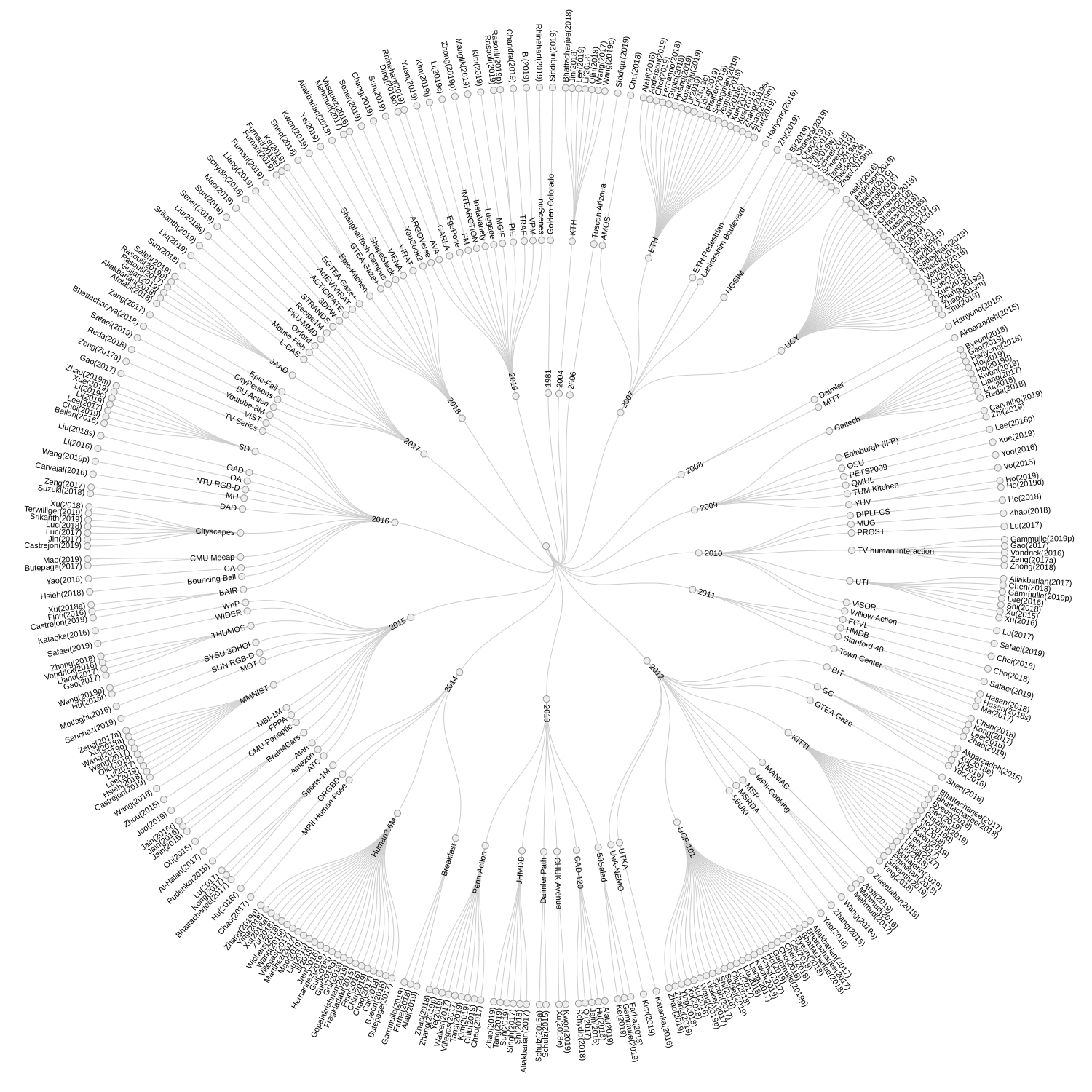}
\caption{An illustration of datasets and papers that use them.}
\label{datasets_fig}
\end{figure*}

We have identified more than 100 datasets that are used in the vision-based prediction literature. Discussing all datasets in detail is beyond the scope of this paper. We provide a summary of the datasets and their characteristics in Tables \ref{dataset_table1} and \ref{dataset_table2} and briefly discuss more popular datasets in each field. Figure \ref{datasets_fig} illustrates the list of papers and corresponding datasets used for evaluation. Note that the papers that do not use publicly available datasets are not listed in this figure. For further information, the readers can also refer to Appendices \ref{links_datasets} and \ref{papers_datasets}. 

\textbf{Video prediction.} Almost any forms of  sequential RGB images can be used for evaluation of video prediction algorithms. Among the most common datasets are traffic datasets such a KITTI \cite{Geiger_2012_CVPR}, and Caltech Pedestrians \cite{Dollar_2009_CVPR}. KITTI is a dataset recorded from inside of a vehicle and contains images of urban roads annotated with bounding box information. It also contains depth maps, LIDAR point clouds and semantic segmentation maps. Caltech Pedestrian is a similar dataset with the difference of only containing RGB images and bounding boxes for pedestrians. It also contains occlusion bounding boxes highlighting the visible portions of the pedestrians. Activity datasets such as UCF-101 \cite{Soomro_2012_arxiv} and Human3.6M \cite{Ionescu_2014_PAMI} are also widely used. UCF-101 contains videos of various types of activities such as sports, applying makeup, playing music instruments annotated with activity labels per video. Human3.6M consists of 3.6 million 3D human poses and corresponding images recorded from 11 professional actors. This dataset contains 17 generic scenarios such as discussion, smoking, and taking photos.

\textbf{Action prediction. } The algorithms in this domain are evaluated on a wide range of datasets. For anticipation tasks, traffic datasets such as Next Generation Simulation (NGSIM) \cite{NGSIM_2007} and Joint Attention in Autonomous Driving (JAAD) \cite{Rasouli_2017_ICCVW} are used.  NGSIM contains trajectories of vehicles driving on highways in the United States. The trajectories are accompanied by the top-down views of the corresponding road structures.   The JAAD dataset contains videos of pedestrians crossing the road recorded using an on-board camera. This dataset contains the frame-wise pedestrian bounding boxes, and action labels as well as pedestrians' and roads' attributes. A similar dataset to JAAD is Pedestrian Intention Estimation (PIE) \cite{Rasouli_2019_ICCV} which, in addition, provides the ego-vehicle sensor data and spatial annotations for traffic elements.

Another popular category of datasets in this domain is those containing videos of cooking activities. These datasets are  Epic-Kitchen \cite{Damen_2018_ECCV}, 50salads \cite{Stein_2013_IJCPUC}, Breakfast \cite{Kuehne_2014_CVPR} and MPII-Cooking \cite{Rohrbach_2012_CVPR}. These datasets contain videos showing sequences of different cooking actions of preparing meals. All videos in the datasets have temporal segments with corresponding activity labels. Some datasets also provide additional annotations such as object bounding boxes, voice and text in  Epic-Kitchen, and the poses of the actors in MPII-Cooking. 

Early action prediction works widely use the popular UCF-101 dataset \cite{Soomro_2012_arxiv} and interaction datasets such as UT Interaction (UTI) \cite{Ryoo_2010_UT} and BIT \cite{Kong_2012_ECCV}. UTI and BIT contain videos of people engaged in interaction with the corresponding label for the types of interactions. In addition, UTI has the added temporal segment annotations detailing different stages of interactions.

\textbf{Trajectory prediction. } The most common datasets in this domain are ETH \cite{Pellegrini_2009_ICCV} and UCY \cite{Lerner_2007_CGF} which contain surveillance videos of pedestrians walking on sidewalks annotated with their position coordinates. UCY also provides the gaze directions to capture the viewing angle of pedestrians. Another popular dataset is Stanford Aerial Pedestrian (SAP), also known as Stanford Drone (SD) \cite{Robicquet_2016_ECCV}. This dataset has the footage of road users from a top-down view recorded by a drone. The annotations include bounding boxes and object class labels. 

\textbf{Motion prediction. } The algorithms in this domain are mainly evaluated on the widely popular dataset Human 3.6M \cite{Ionescu_2014_PAMI} described earlier. This dataset is particularly suitable for these applications because it contains accurate 3D poses of the actors recorded by a high-speed motion capture system. Using this dataset, the background  can be accurately removed allowing the algorithms to focus purely on changes in the poses. Another popular dataset in this field is Penn Action \cite{Zhang_2013_ICCV} which contains RGB videos of various activities with corresponding activity labels and poses of the actors involved. 

\textbf{Other applications. } The most notable datasets are KITTI \cite{Geiger_2012_CVPR} which is used by the OGM prediction algorithms and CityScapes \cite{Cordts_2016_CVPR} that is used by the segmentation prediction algorithms. CityScapes contains video footage of urban environments recorded by an on-board camera. The data is annotated with semantic masks of the traffic objects and corresponding bounding boxes.





\section{Summary and Discussion}
\subsection{Architecture}
There are many factors that define the choice of architecture for vision-based prediction tasks. These factors include the types of input data and expected output, computational efficiency, application-specific constrains, etc.  For instance, in terms of network choice, whether it is feedforward and recurrent, no preference is observed in video applications. However, in the case of action, trajectory and motion predictions, recurrent architectures are strongly preferred. This can be due to the fact that these applications often rely on multi-modal data which can be combined easier in a recurrent framework. In the case of trajectory prediction, recurrent architectures give the flexibility of varying observation and prediction lengths without the need for architectural modifications.  

Generative Adversarial Networks (GANs) are widely used in video prediction applications and to some extent in trajectory prediction methods. Some of the major challenges using generative models is to deal with inherent uncertainty of future representations, in particular, this an issue in the context of trajectory prediction due to high unpredictability of human movement. To remedy this issue and to capture uncertainty of movement, techniques such as variational auto encoders, in which uncertainty is modeled as a latent distribution, and the use of probabilistic objective functions are common.  

A more recent trend in the field of vision-based prediction (and perhaps in other computer vision applications) is the use of attention modules. These modules can be applied at spatial or temporal level or even to adjust the impact of different modalities of data.

\subsection{Data representation and processing}
The type of data and methods of processing vary across different applications. For instance, video applications mainly rely on images but also take advantage of alternative representations, such as optical flow, poses, object-based keypoints, and report improved results. Similarly, many action prediction algorithms use different sources of information such as optical flow, poses, scene attributes, text, complementary sensor readings (e.g. speed in vehicles), gaze and time of the actions.

Trajectory prediction algorithms, predominantly rely on trajectory information, with some exceptions that use scene layouts, complimentary sensors' readings or other constrains. One of the main applications in this domain, in particular surveillance, is modeling the social interaction between the dynamic agents. Unlike other vision-based applications, motion prediction algorithms are mainly single-modal and  use only poses and the images of agents as inputs.

\subsection{Evaluation}
\subsubsection{Metrics}
Metrics may vary across different applications of vision-based prediction. Video prediction algorithms, for instance, are mainly evaluated using  MSE, SSIM, and PSNR, whereas in the case of action prediction algorithms the main metrics are  accuracy, precision, and recall. Trajectory prediction works often measure the average distance (ADE) or final distance (FDE) between the actual and predicted locations of the agents. The models with probabilistic outputs are also evaluated using NLL and KLD metrics. Distance-based metrics are used in motion prediction methods where the error in joint prediction is either calculated on average (MJE) or per joint (MPJPE). In addition, joint accuracy can be reported in terms of the percentage of correct prediction using PCK metric. In this case, a tolerance threshold is defined to determine whether a predicted joint is correct.

While calculating performance error for video and action prediction algorithms are fairly standardized, there are major discrepancies across different works in the way error is computed for trajectory and motion prediction algorithms. For example, in trajectory prediction, distance error is calculated by using metrics such as MSE, RMSE, ED, etc. and units such as pixels and meters. Such discrepancy, and the fact that many works omit mentioning the choice of error metrics and units, increases the chance of incorrect comparisons between models.

\subsubsection{Datasets}
The choice of datasets depends on the objective of the applications. For example, action prediction algorithms for cooking activities are evaluated on datasets such as Epic-Kitchen, 50Salads, Breakfast, and MPII-Cooking and the ones for traffic events evaluated on JAAD, NSGIM, and PIE. Similarly, trajectory prediction works for surveillance widely use UCY, ETH, and SD and for traffic NGSIM. Motion prediction algorithms are more focusing on individual movements in diverse context, therefore predominantly use Human3.6M and Penn Action datasets. 

Compared to other applications, video prediction is an exception. The algorithms in this group are evaluated on almost any datasets with video content. The algorithms in this domain are often task agnostic meaning that the same approaches are evaluated on datasets with traffic scenes (e.g. KITTI, Caltech Pedestrian), general activities (e.g. UCF-101, Penn Action), basic actions (e.g. Human3.6M, KTH) and synthetic data (e.g. MMNIST, Atari games). Although such generalizability across different domains is a desirable feature in video prediction algorithms, it is often the case that the reason behind the choice of datasets is not discussed raising the question of whether the decision for selecting particular datasets is motivated by the limitations of the algorithms.

\subsection{What's next}
In recent years we have witnessed an exponential growth in the number of works published in the field of vision-based prediction. There are still, however, many open research problems in the field that need to be addressed. 

The ability to hallucinate or generate parts of the image that were not previously observed is still a major challenge in video prediction applications. In addition, the algorithms in this domain cannot deal with cases where some objects go out of view in future time frames. The performances of action prediction algorithms are still sub-optimal, especially in safety critical and complex tasks such as event prediction in traffic scenes. To make predictions in such cases, many modalities of data and the relationships between them should be considered which is often not the case in the proposed approaches.

Trajectory prediction algorithms mainly rely on changes in the location of the agents to predict their future states. Although, this might be an effective approach for tasks such as surveillance, in many other cases it might not be sufficient. For example, in order to predict trajectories of pedestrians in traffic scenes, many other sources of information, such as their poses and orientation, road structure, interactions, road conditions, traffic flow, etc., are potentially relevant. Such contextual analysis can also be beneficial for motion prediction algorithms which manly rely on the changes in poses to  predict the future.

In terms of the choice of learning architectures and training schemes, a systematic comparison of different approaches, e.g. using feedforward vs recurrent networks, the benefits of using adversarial training schemes, various uncertainty modeling approaches, etc. is missing. Such information can be partially extracted from the existing literature, however, in many cases it is not possible due to the lack of standard evaluation procedures and metrics, unavailability of corresponding implementation code and the datasets used for comparison.

\ifCLASSOPTIONcaptionsoff
  \newpage
\fi

\bibliographystyle{IEEEtran}
\bibliography{vision_based_prediction}

\appendices
\section{Papers with code}
\label{papers_with_code}
\begin{table*}[]
\centering
\resizebox{1\textwidth}{!}{%
\begin{tabular}{|c|l|l|}
\hline
\textbf{App} & \multicolumn{1}{c|}{\textbf{Paper}} & \multicolumn{1}{c|}{\textbf{Link}} \\ \hline
\multirow{14}{*}{\textbf{Video}}
 & \cite{Wang_2019_BMVC} & \url{https://github.com/andrewjywang/SEENet} \\ \cline{2-3} 
 & \cite{Li_2018_ECCV} & \url{https://github.com/Yijunmaverick/FlowGrounded-VideoPrediction} \\ \cline{2-3} 
 & \cite{Liu_2018_ECCV} & \url{https://github.com/liuem607/DYAN} \\ \cline{2-3} 
 & \cite{Zhao_2018_ECCV} & \url{https://github.com/garyzhao/FRGAN} \\ \cline{2-3} 
 & \cite{Hsieh_2018_NIPS} & \url{https://github.com/jthsieh/DDPAE-video-prediction} \\ \cline{2-3} 
 & \cite{Xu_2018_NIPS} & \url{https://github.com/xjwxjw/VPSS} \\ \cline{2-3} 
 & \cite{Jin_2018_IROS} & \url{https://github.com/jinbeibei/VarNet} \\
\hline
\end{tabular}
}
\caption{A summary of vision-based prediction papers with published code.}
\label{table_published_code}
\end{table*}

A list of papers with official published code can be found in Table \ref{table_published_code}.
\section{Metrics and corresponding papers}
\label{metrics_papers}
\begin{table*}[ht]
\centering
\begin{tabular}{|l|p{0.8\textwidth}|}
\hline
\textbf{Metric} & \textbf{Papers}                                                                                                                                                                                                                                                                                                                                                                                                                                                                                                                                                                                                                                                                                                                                                                                                                                                                                                                                                                                                                                                                                                                \\
\hline
\end{tabular}
\caption{Metrics used in \textbf{video prediction} applications.}
\label{video_metrics_table}
\end{table*}

\begin{table*}[ht]
\centering
\begin{tabular}{|l|p{0.8\textwidth}|}
\hline
\textbf{Metrics}            & \textbf{Papers}                                                                                                                                                                                                                                                                                                                                                                                                                                                                                                                                                                                                                                                                                                                                                                                                                                                                                                                                                                                                                                                                                                                                                                                                                                                                                                                                                                                                                                                                                                                                                                                                                                                                                                                                                                                                                                                                                                                                                                                                                                                                                                                                                                                                                                                                                                                                                                                                                                                                                                                                                                                   \\
\hline
\end{tabular}
\caption{Metrics used in \textbf{action prediction} applications.}
\label{action_metrics_table}
\end{table*}

\begin{table*}[ht]
\centering
\begin{tabular}{|l|p{0.8\textwidth}|}
\hline
\textbf{Metrics}            & \textbf{Papers}                                                                                                                                                                                                                                                                                                                                                                                                                                                                                                                                                                                                                                                                                                                                                                                                                                                                                                                                                                                                                                                                                                                                                                                                                                                                                                                                                                                                                                                                                                                                                                                                                                                                                                                                                                                                                                                                                                                                                                                                                                                                                                                        \\
\hline
\end{tabular}
\caption{Metrics used in \textbf{trajectory prediction} applications.}
\label{trajectory_metrics_table}
\end{table*}

\begin{table*}[h]
\centering
\begin{tabular}{|l|l|}
\hline
\textbf{Metrics}           & \textbf{Papers}                                                                                                                                                                                                                                                                                                                                                                                                                                                             \\ \hline
\textit{\textbf{Accuracy}} & \cite{Luo_2019_IROS},\cite{Fragkiadaki_2015_ICCV}                                                                                                                                                                                                                                                                                                                                                                                     \\ \hline
\textit{\textbf{Human}}    & \cite{Hernandez_2019_ICCV},\cite{Gui_2018_ECCV_2}                                                                                                                                                                                                                                                                                                                                                                                    \\ \hline
\textit{\textbf{LO}}       & \cite{Joo_2019_CVPR}                                                                                                                                                                                                                                                                                                                                                                                                                                     \\
\hline
\end{tabular}
\caption{Metrics used in motion prediction applications.}
\label{motion_metrics_table}
\end{table*}

Lists of metrics and corresponding papers can be found in Tables \ref{video_metrics_table}, \ref{action_metrics_table}, \ref{trajectory_metrics_table}, \ref{motion_metrics_table}, and \ref{other_metrics_table}. 

\begin{table*}[]
\centering
\begin{tabular}{|l|l|}
\hline
\textbf{Metrics}            & \textbf{Papers}                                                                                                                   \\ \hline
\textit{\textbf{AUC}}       & \cite{Choi_2016_IROS}                                                                                          \\ \hline
\textit{\textbf{Accuracy}}  & \cite{Zeng_2017_ICCV},\cite{Carvajal_2016_ICPR},\cite{Joo_2015_ICCV}     \\ \hline
\textit{\textbf{ED}}        & \cite{Graves_2019_IROS},\cite{Zapf_2019_IROS}                                               \\ \hline
\textit{\textbf{EPE}}       & \cite{Jin_2017_NIPS}                                                                                           \\ \hline
\textit{\textbf{F1}}        & \cite{Guizilini_2019_ICRA},\cite{Schreiber_2019_ICRA}                                       \\ \hline
\textit{\textbf{GCE}}       & \cite{Luc_2018_ECCV}                                                                                           \\ \hline
\textit{\textbf{ISM}}       & \cite{Afolabi_2018_IROS}                                                                                       \\
\hline
\end{tabular}
\caption{Metrics used in \textbf{other prediction} applications.}
\label{other_metrics_table}
\end{table*}

\clearpage

\clearpage
\section{Metric formulas}
\label{appendix_metrics}

\subsection{Video prediction}

 $$ MSE= \frac{1}{MN} \sum_{i=1}^M\sum_{j=1}^N(I(i,j) -\tilde{I}(i,j))^2$$

 $$PSNR = 20\log\left(\frac{MAX_I}{\sqrt{MSE}}\right)$$
 
\textbf{Structural Similarity (SSIM)} 
$$luminance (l)(x,y) = \frac{2\mu_x\mu_y + C_1}{\mu_x^2 + \mu_y^2 + C_1}$$

$$\mu_x=\frac{1}{N}\sum_{i=1}^Nx_i$$

$$C_1=(K_1L)^2$$

\noindent where $L$ is dynamic range of pixel values (e.g. 255) and $K_1 \ll 1$ is a small constant.

$$contrast (c)(x,y) = \frac{2\sigma_x\sigma_y+C_2}{\sigma_x^2+\sigma_y^2+C_2}$$

$$\sigma_x=\left(\frac{1}{N-1}\sum_{i=1}^N(x_i-\mu_x)^2\right)^{1/2}$$

$$C_2=(K_2L)^2$$ $$K_2 \ll 1$$

$$structure(s)(x,y) = \frac{\sigma_{xy}+C_3}{\sigma_x\sigma_y+C_3}$$

$$C_3=(K_3L)^2$$ $$K_3 \ll 1$$

$$SSIM(x,y)=[l(x,y)]^\alpha.[c(x,y)]^\beta.[s(x,y)]^\gamma$$

\noindent where $\alpha ,\beta, \gamma >0$ are parameters to choose in order to adjust the importance.

\textbf{Learned Perceptual Image Patch Similarity (LPIPS)}

Assume features are extracted from $L$ layers and unit-normalized in channel dimension, for layer l

$$\hat{y}^l,\hat{y}^l \in R^{H_l\times W_l \times C_l}$$.

The distance between reference $x$ and distorted patches $x_0$ is given by,

$$d(x,x_0)=\sum_l \frac{1}{H_lW_l}\sum_{w,l}\parallel w_l \odot (\hat{y}^l_{hw},\hat{y}^l_{0hw})\parallel ^2_2$$  

\subsection{Action prediction}
There are 4 possibilities for classification: True positive (TP) and True Negative (TN) when the algorithm correctly classifies positive and negative samples, and False Positive (FP) and False Negative (FN) when the algorithm incorrectly classifies negative samples as positive and vice versa.

$$Accuracy = \frac{TN+TP}{TP+TN+FP+FN}$$
$$Precision = \frac{TP}{TP+FP}$$
$$Recall = \frac{TP}{TP+FN}$$
$$F1-score = 2 \times \frac{Precision \times Recall}{Precision + Recall}$$

$$RMSE = \sqrt{\frac{1}{n} \sum_{i=1}^n(y_i -\tilde{y_i})}$$

Let $p(r)$ be the precision-recall curve. Then,

$$AP = \int_0^1 p(r)dr$$

\subsection{Trajectory prediction}

\subsubsection{Distance metrics}
$$\text{Euclidean Distance}(ED) = \parallel y-\tilde{y}\parallel = \parallel y-\tilde{y}\parallel_2$$
$$ = \sqrt{\sum_{i=1}^n(y_i -\tilde{y_i})^2} $$
$$ \text{Mean Absolute Error}(MAE) = \frac{1}{n} \sum_{i=1}^n|y_i -\tilde{y_i}| $$
$$\text{Mean Square Error} (MSE) = \parallel y-\tilde{y}\parallel^2$$
$$ =\frac{1}{n} \sum_{i=1}^n(y_i -\tilde{y_i})^2$$
$$\text{Root MSE} (RMSE) = \sqrt{\frac{1}{n} \sum_{i=1}^n(y_i -\tilde{y_i})^2}$$
$$\text{Hausdorff Distance}(HD) = max_{y\in Y}min_{\tilde{y}\in\tilde{Y}} \parallel y-\tilde{y}\parallel$$
$$\text{Modified HD}(MHD) = max(d(Y,\tilde{Y}),d(\tilde{Y}, Y))$$
$$ d(Y,\tilde{Y}) = \frac{1}{N_y}\sum_{y\in Y} min_{\tilde{y}\in \tilde{Y}} \parallel y - \tilde{y}\parallel$$

$$ ADE = \frac{\sum^N_{i=1}\sum^{T_{pred}}_{t=1}\parallel \tilde{y}^i_t - y^i_t \parallel}{N \times T_{pred}}$$
where $N$ is the number of samples and $T_{pred}$ is the prediction steps.

$$FDE = \frac{ \sum^N_{i=1} \parallel \tilde{y}^i_{T_{pred}} - y^i_{T_{pred}} \parallel}{N}$$

$$ minMSD = \mathbb{E}_{\tilde{Y}_k\sim q_\theta}min_{\tilde{y}\in\tilde{Y}_k}\parallel y-\tilde{y}\parallel^2$$
where $q_\theta$ is the sampling space and $K$ number of samples.

$$ meanMSD = \frac{1}{K}\sum_{k=1}^K\parallel y-\tilde{y}\parallel^2$$

$$NLL = \mathbb{E}_{p(Y|X)}\left[-\log \prod_{t=1}^{T_{pred}} p(y_t|X)\right]$$

\subsection{Motion prediction}

$$MPJPE = \frac{1}{N \times T_{pred}}\sum_{t=1}^{T_{pred}}\sum_{i=1}^N \parallel (J_i^t-J_{root}^t)-(\tilde{J}_i^t-\tilde{J}_{root}^t)\parallel$$

\clearpage
\section{Links to the datasets}
\label{links_datasets}

\begin{table*}[]
\centering
\resizebox{1\textwidth}{!}{%
\begin{tabular}{|l|l|l|}
\hline
Year & Dataset & Links \\ \hline
\multirow{12}{*}{2019} & ARGOVerse\cite{Chang_2019_CVPR} & \url{https://www.argoverse.org/data.html} \\ \cline{2-3} 
 & CARLA\cite{Rhinehart_2019_ICCV} & \url{https://sites.google.com/view/precog} \\ \cline{2-3} 
 & EgoPose\cite{Yuan_2019_ICCV} & \url{https://github.com/Khrylx/EgoPose} \\ \cline{2-3} 
 & FM\cite{Kim_2019_ICCV} & \url{https://mcl.korea.ac.kr/$\sim$krkim/iccv2019/index.html} \\ \cline{2-3} 
 & InstaVariety\cite{Kanazawa_2019_CVPR} & \url{https://github.com/akanazawa/human_dynamics} \\ \cline{2-3} 
 & INTEARCTION\cite{Zhan_2019_arxiv} & \url{https://interaction-dataset.com} \\ \cline{2-3} 
 & Luggage\cite{Manglik_2019_IROS} & \url{https://aashi7.github.io/NearCollision.html} \\ \cline{2-3} 
 & MGIF\cite{Siarohin_2019_CVPR} & \url{https://github.com/AliaksandrSiarohin/monkey-net} \\ \cline{2-3} 
 & PIE\cite{Rasouli_2019_ICCV} & \url{http://data.nvision2.eecs.yorku.ca/PIE_dataset/} \\ \cline{2-3} 
 & nuScenes\cite{Caesar_2019_arxiv} & \url{https://www.nuscenes.org/} \\ \cline{2-3} 
 & Vehicle-Pedestrian-Mixed (VPM)\cite{Bi_2019_ICCV} & \url{http://vr.ict.ac.cn/vp-lstm.} \\ \cline{2-3} 
 & TRAF\cite{Chandra_2019_CVPR} & \url{https://drive.google.com/drive/folders/1LqzJuRkx5yhOcjWFORO5WZ97v6jg8RHN} \\ \hline
\multirow{10}{*}{2018} & 3DPW\cite{vonMarcard_2018_ECCV} & \url{https://virtualhumans.mpi-inf.mpg.de/3DPW/} \\ \cline{2-3} 
 & ActEV/VIRAT\cite{Awad_2018_Trecvid} & \url{https://actev.nist.gov/trecvid19} \\ \cline{2-3} 
 & ACTICIPATE\cite{Schydlo_2018_ICRA} & \url{http://vislab.isr.tecnico.ulisboa.pt/datasets/} \\ \cline{2-3} 
 & AVA\cite{Gu_2018_CVPR} & \url{https://research.google.com/ava/} \\ \cline{2-3} 
 & Epic-Kitchen\cite{Damen_2018_ECCV} & \url{https://epic-kitchens.github.io/2019} \\ \cline{2-3} 
 & EGTEA Gaze+\cite{Li_2018_ECCV_2} & \url{http://www.cbi.gatech.edu/fpv/} \\ \cline{2-3} 
 & STC\cite{Liu_2018_CVPR} & \url{https://svip-lab.github.io/dataset/campus_dataset.html} \\ \cline{2-3} 
 & ShapeStack\cite{Groth_2018_arxiv} & \url{https://shapestacks.robots.ox.ac.uk/} \\ \cline{2-3} 
 & VIENA\cite{Aliakbarian_2018_ACCV} & \url{https://sites.google.com/view/viena2-project/home} \\ \cline{2-3} 
 & YouCook2\cite{Zhou_2018_AI} & \url{http://youcook2.eecs.umich.edu/} \\ \hline
\multirow{10}{*}{2017} & BUA\cite{Ma_2017_PR} & \url{http://cs-people.bu.edu/sbargal/BU-action/} \\ \cline{2-3} 
 & CityPerson\cite{Shanshan_2017_CVPR} & \url{https://bitbucket.org/shanshanzhang/citypersons/src/default/} \\ \cline{2-3} 
 & Epic-fail\cite{Zeng_2017_CVPR} & \url{http://aliensunmin.github.io/project/video-Forecasting/} \\ \cline{2-3} 
 & JAAD\cite{Rasouli_2017_ICCVW} & \url{http://data.nvision2.eecs.yorku.ca/JAAD_dataset/} \\ \cline{2-3} 
 & L-CAS\cite{Yan_2017_IROS} & \url{https://lcas.lincoln.ac.uk/wp/research/data-sets-software/l-cas-3d-point-cloud-people-dataset/} \\ \cline{2-3} 
 & Mouse Fish \cite{Xu_2017_IJCV} & \url{https://web.bii.a-star.edu.sg/archive/machine_learning/Projects/behaviorAnalysis/Lie-X/Lie-X.html} \\ \cline{2-3} 
 & ORC\cite{Maddern_2017_IJRR} & \url{https://robotcar-dataset.robots.ox.ac.uk/} \\ \cline{2-3} 
 & PKU-MMD\cite{Liu_2017_arxiv} & \url{http://www.icst.pku.edu.cn/struct/Projects/PKUMMD.html} \\ \cline{2-3} 
 & Recipe1M\cite{Salvador_2017_CVPR} & \url{http://pic2recipe.csail.mit.edu/} \\ \cline{2-3} 
 & STRANDS\cite{Hawes_2017_RAM} & \url{https://strands.readthedocs.io/en/latest/datasets/} \\ \hline
\multirow{13}{*}{2016} & BAIR Push\cite{Finn_2016_NIPS} & \url{https://sites.google.com/site/brainrobotdata/home/push-dataset} \\ \cline{2-3} 
 & BB\cite{Chang_2016_arxiv} & \url{https://github.com/mbchang/dynamics} \\ \cline{2-3} 
 & MU\cite{Carvajal_2016_ICPR} & \url{http://staff.itee.uq.edu.au/lovell/MissUniverse/} \\ \cline{2-3} 
 & Cityscapes\cite{Cordts_2016_CVPR} & \url{https://www.cityscapes-dataset.com/} \\ \cline{2-3} 
 & CMU mocap\cite{CMU_Mocap_2016} & \url{http://mocap.cs.cmu.edu/} \\ \cline{2-3} 
 & DAD\cite{Chan_2016_ACCV} & \url{https://aliensunmin.github.io/project/dashcam/} \\ \cline{2-3} 
 & NTU RGB-D\cite{Shahroudy_2016_CVPR} & \url{http://rose1.ntu.edu.sg/Datasets/actionRecognition.asp} \\ \cline{2-3} 
 & OA\cite{Li_2016_WACV} & \url{http://www.mpii.de/ongoing-activity} \\ \cline{2-3} 
 & OAD\cite{Li_2016_ECCV} & \url{http://www.icst.pku.edu.cn/struct/Projects/OAD.html} \\ \cline{2-3} 
 & SD\cite{Robicquet_2016_ECCV} & \url{http://cvgl.stanford.edu/projects/uav_data/} \\ \cline{2-3} 
 & TV Series\cite{De_2016_ECCV} & \url{https://github.com/zhenyangli/online_action} \\ \cline{2-3} 
 & VIST\cite{Huang_2016_NAACL} & \url{http://visionandlanguage.net/VIST/} \\ \cline{2-3} 
 & Youtube-8M\cite{Abu_2016_arxiv} & \url{https://research.google.com/youtube8m/} \\ \hline
\multirow{14}{*}{2015} & Amazon\cite{Mcauley_2015_CRDIR} & \url{http://jmcauley.ucsd.edu/data/amazon/index_2014.html} \\ \cline{2-3} 
 & Atari\cite{Oh_2015_NIPS} & \url{https://github.com/junhyukoh/nips2015-action-conditional-video-prediction} \\ \cline{2-3} 
 & Brain4Cars\cite{Jain_2015_ICCV} & \url{https://github.com/asheshjain399/ICCV2015_Brain4Cars} \\ \cline{2-3} 
 & CMU Panoptic\cite{Joo_2015_ICCV_2} & \url{http://domedb.perception.cs.cmu.edu/dataset.html} \\ \cline{2-3} 
 & FPPA\cite{Zhou_2015_ICCV} & \url{http://bvision11.cs.unc.edu/bigpen/yipin/ICCV2015/prediction_webpage/Prediction.html} \\ \cline{2-3} 
 & GTEA Gaze+\cite{Li_2015_CVPR} & \url{http://www.cbi.gatech.edu/fpv/} \\ \cline{2-3} 
 & MBI-1M\cite{Cappallo_2015_ICMR} & \url{http://academic.mywebsiteontheinternet.com/data/} \\ \cline{2-3} 
 & MOT\cite{Leal_2015_arxiv} & \url{https://motchallenge.net/} \\ \cline{2-3} 
 & MMNIST\cite{Srivastava_2015_ICML} & \url{http://www.cs.toronto.edu/$\sim$nitish/unsupervised_video/} \\ \cline{2-3} 
 & SUN RGB-D\cite{Song_2015_CVPR_2} & \url{http://rgbd.cs.princeton.edu/} \\ \cline{2-3} 
 & SYSU 3DHOI\cite{Hu_2015_CVPR} & \url{http://www.isee-ai.cn/$\sim$hujianfang/ProjectJOULE.html} \\ \cline{2-3} 
 & THUMOS\cite{Gorban_2015} & \url{http://www.thumos.info/home.html} \\ \cline{2-3} 
 & WnP\cite{Wu_2015_CVPR} & \url{http://watchnpatch.cs.cornell.edu/} \\ \cline{2-3} 
 & Wider\cite{Xiong_2015_CVPR} & \url{http://yjxiong.me/event_recog/WIDER/} \\ \hline
\multicolumn{1}{|c|}{\multirow{5}{*}{2014}} & Breakfast\cite{Kuehne_2014_CVPR} & \url{http://serre-lab.clps.brown.edu/resource/breakfast-actions-dataset/} \\ \cline{2-3} 
\multicolumn{1}{|c|}{} & Human3.6M\cite{Ionescu_2014_PAMI} & \url{http://vision.imar.ro/human3.6m/description.php} \\ \cline{2-3} 
\multicolumn{1}{|c|}{} & MPII Human Pose\cite{Andriluka_2014_CVPR} & \url{http://human-pose.mpi-inf.mpg.de/} \\ \cline{2-3} 
\multicolumn{1}{|c|}{} & ORGBD\cite{Yu_2014_ACCV} & \url{https://sites.google.com/site/skicyyu/orgbd} \\ \cline{2-3} 
\multicolumn{1}{|c|}{} & Sports-1M\cite{Karpathy_2014_CVPR} & \url{https://cs.stanford.edu/people/karpathy/deepvideo/} \\ \hline
\end{tabular}
}
\caption{A summary of datasets (from year 2014-2019)  used in vision-based prediction papers and corresponding links.}

\label{data_set_links1}
\end{table*}

\begin{table*}[]
\centering
\resizebox{1\textwidth}{!}{%
\begin{tabular}{|l|l|l|}
\hline
Year & Dataset & Links \\ \hline
\multirow{7}{*}{2013} & 50 salads\cite{Stein_2013_IJCPUC} & \url{https://cvip.computing.dundee.ac.uk/datasets/foodpreparation/50salads/} \\ \cline{2-3} 
 & ATC \cite{Brvsvcic_2013_HMS} & \url{https://irc.atr.jp/crest2010_HRI/ATC_dataset/} \\ \cline{2-3} 
 & CAD-120\cite{Koppula_2013_IJRR} & \url{http://pr.cs.cornell.edu/humanactivities/data.php} \\ \cline{2-3} 
 & CHUK Avenue\cite{Lu_2013_ICCV} & \url{http://www.cse.cuhk.edu.hk/leojia/projects/detectabnormal/dataset.html} \\ \cline{2-3} 
 & Daimler path\cite{Schneider_2013_GCPR} & \url{http://www.gavrila.net/Datasets/Daimler_Pedestrian_Benchmark_D/Pedestrian_Path_Predict_GCPR_1/pedestrian_path_predict_gcpr_1.html} \\ \cline{2-3} 
 & JHMDB\cite{Jhuang_2013_ICCV} & \url{http://jhmdb.is.tue.mpg.de/} \\ \cline{2-3} 
 & Penn Action\cite{Zhang_2013_ICCV} & \url{http://dreamdragon.github.io/PennAction/} \\ \hline
\multirow{11}{*}{2012} & BIT\cite{Kong_2012_ECCV} & \url{https://sites.google.com/site/alexkongy/software} \\ \cline{2-3} 
 & GTEA Gaze\cite{Fathi_2012_ECCV} & \url{http://www.cbi.gatech.edu/fpv/} \\ \cline{2-3} 
 & KITTI\cite{Geiger_2012_CVPR} & \url{http://www.cvlibs.net/datasets/kitti/} \\ \cline{2-3} 
 & MANIAC\cite{Abramov_2012_WACV} & \url{https://alexandria.physik3.uni-goettingen.de/cns-group/datasets/maniac/} \\ \cline{2-3} 
 & MPII-cooking\cite{Rohrbach_2012_CVPR} & \url{https://www.mpi-inf.mpg.de/departments/computer-vision-and-machine-learning/research/human-activity-recognition/mpii-cooking-activities-dataset/} \\ \cline{2-3} 
 & MSRDA\cite{Wang_2012_CVPR} & \url{https://documents.uow.edu.au/$\sim$wanqing/\#MSRAction3DDatasets} \\ \cline{2-3} 
 & GC\cite{Zhou_2012_CVPR} & \url{http://www.ee.cuhk.edu.hk/$\sim$xgwang/grandcentral.html} \\ \cline{2-3} 
 & SBUKI\cite{kiwon_2012_CVPR} & \url{https://www3.cs.stonybrook.edu/$\sim$kyun/research/kinect_interaction/index.html} \\ \cline{2-3} 
 & UCF-101\cite{Soomro_2012_arxiv} & \url{https://www.crcv.ucf.edu/data/UCF101.php} \\ \cline{2-3} 
 & UTKA\cite{Xia_2012_CVPRW} & \url{http://cvrc.ece.utexas.edu/KinectDatasets/HOJ3D.html} \\ \cline{2-3} 
 & UvA-NEMO\cite{Dibeklio_2012_ECCV} & \url{https://www.uva-nemo.org/} \\ \hline
\multirow{5}{*}{2011} & FCVL\cite{Pandey_2011_IJRR} & \url{http://robots.engin.umich.edu/SoftwareData/Ford} \\ \cline{2-3} 
 & HMDB\cite{Kuehne_2011_ICCV} & \url{http://serre-lab.clps.brown.edu/resource/hmdb-a-large-human-motion-database/} \\ \cline{2-3} 
 & Stanford 40\cite{Yao_2011_ICCV} & \url{http://vision.stanford.edu/Datasets/40actions.html} \\ \cline{2-3} 
 & Town Center\cite{Benfold_2011_CVPR} & \url{http://www.robots.ox.ac.uk/ActiveVision/Research/Projects/2009bbenfold_headpose/project.html\#datasets} \\ \cline{2-3} 
 & VIRAT\cite{Oh_2011_CVPR} & \url{http://viratdata.org/} \\ \hline
\multirow{8}{*}{2010} & DISPLECS\cite{Pugeault_2010_ECCV} & \url{https://cvssp.org/data/diplecs/} \\ \cline{2-3} 
 & MSR\cite{Li_2010_CVPRW} & \url{https://www.microsoft.com/en-us/download/details.aspx?id=52315} \\ \cline{2-3} 
 & MUG Facial Expression\cite{Aifanti_2010_WIAMIS} & \url{https://mug.ee.auth.gr/fed/} \\ \cline{2-3} 
 & PROST\cite{Santner_2010_CVPR} & \url{www.gpu4vision.com } \\ \cline{2-3} 
 & THI\cite{Patron_2010_BMVC} & \url{http://www.robots.ox.ac.uk/$\sim$alonso/tv_human_interactions.html} \\ \cline{2-3} 
 & UTI\cite{Ryoo_2010_UT} & \url{http://cvrc.ece.utexas.edu/SDHA2010/Human_Interaction.html} \\ \cline{2-3} 
 & VISOR\cite{Vezzani_2010_MTA} & \url{imagelab.ing.unimore.it/visor} \\ \cline{2-3} 
 & Willow Action\cite{Delaitre_2010_BMVC} & \url{https://www.di.ens.fr/willow/research/stillactions/} \\ \hline
\multirow{9}{*}{2009} & Caltech Pedestrian\cite{Dollar_2009_CVPR} & \url{http://www.vision.caltech.edu/Image_Datasets/CaltechPedestrians/} \\ \cline{2-3} 
 & Collective Activity (CA)\cite{Choi_2009_ICCVW} & \url{http://www-personal.umich.edu/$\sim$wgchoi/eccv12/wongun_eccv12.html} \\ \cline{2-3} 
 & EIFP\cite{Majecka_2009} & \url{http://homepages.inf.ed.ac.uk/rbf/FORUMTRACKING/} \\ \cline{2-3} 
 & ETH\cite{Pellegrini_2009_ICCV} & \url{http://www.vision.ee.ethz.ch/en/datasets/} \\ \cline{2-3} 
 & OSU\cite{Hess_2009_CVPR} & \url{http://eecs.oregonstate.edu/football/tracking/dataset} \\ \cline{2-3} 
 & PETS2009\cite{Ferryman_2009_PETS} & \url{http://www.cvg.reading.ac.uk/PETS2009/a.html} \\ \cline{2-3} 
 & QMUL\cite{Loy_2009_BMVC} & \url{http://personal.ie.cuhk.edu.hk/$\sim$ccloy/downloads_qmul_junction.html} \\ \cline{2-3} 
 & TUM Kitchen\cite{Tenorth_2009_ICCVW} & \url{https://ias.in.tum.de/dokuwiki/software/kitchen-activity-data} \\ \cline{2-3} 
 & YUV Videos\cite{ASU_2009_YUV} & \url{http://trace.kom.aau.dk/yuv/index.html} \\ \hline
\multirow{2}{*}{2008} & Daimler\cite{Enzweiler_2008_PAMI} & \url{http://www.gavrila.net/Datasets/Daimler_Pedestrian_Benchmark_D/Daimler_Mono_Ped__Detection_Be/daimler_mono_ped__detection_be.html} \\ \cline{2-3} 
 & MITT\cite{Grimson_2008_CVPR} & \url{http://www.ee.cuhk.edu.hk/$\sim$xgwang/MITtrajsingle.html} \\ \hline
\multirow{5}{*}{2007} & AMOS\cite{Jacobs_2007_CVPR} & \url{http://amos.cse.wustl.edu/} \\ \cline{2-3} 
 & ETH pedestrian\cite{Ess_2007_ICCV} & \url{https://data.vision.ee.ethz.ch/cvl/aess/} \\ \cline{2-3} 
 & Lankershim Boulevard\cite{US_2007_Lankershim} & \url{https://www.fhwa.dot.gov/publications/research/operations/07029/index.cfm} \\ \cline{2-3} 
 & NGSIM\cite{NGSIM_2007} & \url{https://ops.fhwa.dot.gov/trafficanalysistools/ngsim.htm} \\ \cline{2-3} 
 & UCY\cite{Lerner_2007_CGF} & \url{https://graphics.cs.ucy.ac.cy/research/downloads/crowd-data} \\ \hline
2006 & Tuscan Arizona\cite{Pickering_2006} & \url{http://www.mmto.org/} \\ \hline
2004 & KTH\cite{Schuldt_2004_ICPR} & \url{http://www.nada.kth.se/cvap/actions/} \\ \hline
1981 & Golden Colorado\cite{Stoffel_1981} & \url{https://www.osti.gov/dataexplorer/biblio/dataset/1052221} \\ \hline
\end{tabular}
}
\caption{A summary of datasets (from year 2013 and earlier) used in vision-based prediction papers and corresponding links.}
\label{data_set_links2}
\end{table*}

Lists of datasets with associated repository links can be found in Tables \ref{data_set_links1} and \ref{data_set_links2}.

\clearpage
\section{Datasets and corresponding papers}
\label{papers_datasets}
Lists of datasets and corresponding papers can be found in Tables \ref{video_datasets_table}, \ref{action_datasets_table}, \ref{trajectory_datasets_table}, \ref{motion_datasets_table}, and \ref{other_datasets_table}. 

\begin{table*}[]
\centering
\begin{tabular}{|l|l|}
\hline
\textbf{Datasets}                     & \textbf{Papers}                                                                                                                                                                                                                                                                                                                                                                                                                                                                                                                                                                                              \\ \hline
\textit{\textbf{Atari}}               & \cite{Oh_2015_NIPS}                                                                                                                                                                                                                                                                                                                                                                                                                                                                                                                                                                       \\
\hline
\end{tabular}
\caption{Datasets used in \textbf{video prediction} applications.}
\label{video_datasets_table}
\end{table*}

\begin{table*}[]
\centering
\begin{tabular}{|l|l|}
\hline
\textbf{Datasets}                      & \textbf{Papers}                                                                                                                                                                                                                                                                                                                                                                                                                                                                                                                                                                                                                                                                                                                                                                                                                                                                                                                \\ \hline
\textit{\textbf{50Salad}}              & \cite{Ke_2019_CVPR},\cite{Gammulle_2019_BMVC},\cite{Farha_2018_CVPR}                                                                                                                                                                                                                                                                                                                                                                                                                                                                                                                                                                                                                                                                                                                                                                                                  \\
\hline
\end{tabular}
\caption{Datasets used in \textbf{action prediction} applications.}
\label{action_datasets_table}
\end{table*}

\begin{table*}[]
\centering
\begin{tabular}{|l|p{0.8\textwidth}|}
\hline
\textbf{Datasets}                      & \textbf{Papers}                                                                                                                                                                                                                                                                                                                                                                                                                                                                                                                                                                                                                                                                                                                                                                                                                                                                                                                                                                                                                                         \\
\hline
\end{tabular}
\caption{Datasets used in \textbf{trajectory prediction} applications.}
\label{trajectory_datasets_table}
\end{table*}

\begin{table*}[]
\centering
\begin{tabular}{|l|l|}
\hline
\textbf{Datasets}                 & \textbf{Papers}                                                                                                                                                                                                                                                                                                                                                                                                                                                                                                                                                                                                                                                     \\ \hline
\textit{\textbf{3DPW}}            & \cite{Mao_2019_ICCV}                                                                                                                                                                                                                                                                                                                                                                                                                                                                                                                                                                                                                             \\
\hline
\end{tabular}
\caption{Datasets used in \textbf{motion prediction} applications.}
\label{motion_datasets_table}
\end{table*}

\begin{table*}[]
\centering
\begin{tabular}{|l|l|}
\hline
\textbf{Datasets}                 & \textbf{Papers}                                                                                                                                                         \\ \hline
\textit{\textbf{AMOS}}            & \cite{Chu_2018_WACV}                                                                                                                                 \\ \hline
\textit{\textbf{Amazon}}          & \cite{Al-Halah_2017_ICCV}                                                                                                                            \\ \hline
\textit{\textbf{Cityscapes}}      & \cite{Terwilliger_2019_WACV},\cite{Luc_2018_ECCV},\cite{Luc_2017_ICCV},\cite{Jin_2017_NIPS} \\ \hline
\textit{\textbf{DIPLECS}}         & \cite{He_2018_ICPR}                                                                                                                                  \\ \hline
\textit{\textbf{FCVL}}            & \cite{Choi_2016_IROS}                                                                                                                                \\ \hline
\textit{\textbf{Golden Colorado}} & \cite{Siddiqui_2019_WACV}                                                                                                                            \\ \hline
\textit{\textbf{JAAD}}            & \cite{Afolabi_2018_IROS}                                                                                                                             \\
\hline
\end{tabular}
\caption{Datasets used in \textbf{other prediction} applications.}
\label{other_datasets_table}
\end{table*}

\end{document}